\documentclass{article}

\usepackage{microtype}
\usepackage{graphicx}
\usepackage{subfigure}
\usepackage{booktabs} % for professional tables
\usepackage{amsmath}
\usepackage{bm}
\usepackage{amssymb}
\usepackage{graphicx}
\usepackage{wrapfig}
\usepackage{nicefrac}
\usepackage{multirow}
\usepackage{appendix}
\usepackage{hyperref}

\usepackage[accepted]{icml2020}

\icmltitlerunning{Bayesian Structure Adaptation for Continual Learning}

\def\hyphenateAndTtWholeString #1{\xHyphenate#1$\wholeString\unskip}

\def\xHyphenate#1#2\wholeString {\if#1$%
    \else\transform{#1}%
    \takeTheRest#2\ofTheString\fi}

\def\takeTheRest#1\ofTheString\fi
{\fi \xHyphenate#1\wholeString}

\def\transform#1{\url{#1}\hskip 0pt plus 1pt}

\def\urlx #1{\href{#1}{\hyphenateAndTtWholeString{#1}}}

\begin{document}

\twocolumn[
\icmltitle{Bayesian Structure Adaptation for Continual Learning}



\icmlsetsymbol{equal}{*}

\begin{icmlauthorlist}
\icmlauthor{Abhishek Kumar}{equal,to}
\icmlauthor{Sunabha Chatterjee}{equal,to}
\icmlauthor{Piyush Rai}{to}
\end{icmlauthorlist}

\icmlaffiliation{to}{Indian Institute of Technology, Kanpur, Uttar Pradesh, India.}

\icmlcorrespondingauthor{Abhishek Kumar}{abhikcr@iitk.ac.in}
\icmlcorrespondingauthor{Sunabha Chatterjee}{sunabhac@gmail.com}
\icmlcorrespondingauthor{Piyush Rai}{piyush@cse.iitk.ac.in}
% \icmlkeywords{Machine Learning}

\vskip 0.3in
]



\printAffiliationsAndNotice{\icmlEqualContribution} % otherwise use the standard text.
\begin{abstract}
Continual Learning is a learning paradigm where learning systems are trained with sequential or streaming tasks. Two notable directions among the recent advances in continual learning with neural networks are ($i$) variational Bayes based regularization by learning priors from previous tasks, and, ($ii$) learning the structure of deep networks to adapt to new tasks.  So far, these two approaches have been orthogonal. We present a novel Bayesian approach to continual learning based on learning the structure of deep neural networks, addressing the shortcomings of both these approaches. The proposed model learns the deep structure for each task by learning which weights to be used, and supports inter-task transfer through the overlapping of different sparse subsets of weights learned by different tasks. Experimental results on supervised and unsupervised benchmarks shows that our model performs comparably or better than recent advances in continual learning setting.
\end{abstract}
\vspace{-2em}
\section{Introduction}

Continual learning \cite{ring1997child,conti_review} is the learning paradigm where a single model is subjected to a sequence of tasks. At any point of time, the model is expected to ($i$) make predictions for the tasks it has seen so far, ($ii$) if subjected to training data for a new task, adapt to the new task leveraging past experience if possible (forward transfer) and benefit the previous tasks if possible (backward transfer). While the desirable aspects of more mainstream transfer learning (sharing of bias between related tasks \cite{xfer_survey} might reasonably be expected here too, the principal challenge is to retain the predictive power for the older tasks even after learning new tasks, thus avoiding the so-called \textit{catastrophic forgetting}. Real world applications in, for example, robotics or time-series forecasting, are rife with this challenging learning scenario, the ability to adapt to dynamically changing environments or evolving data distributions being essential in these domains. Continual learning is also desirable in unsupervised learning problems as well~\cite{smith2019unsupervised,rao2019continual} where the goal is to learn the underlying structure or latent representation of the data. Also, as a skill innate to humans \cite{continual_human}, it is naturally an interesting scientific problem to reproduce the same capability in artificial predictive modelling systems.

%With the versatility and expressive power of neural networks being well-established in the landscape of machine learning models, current approaches focus on endowing them with the above capability \cite{conti_review}. 
Existing approaches to continual learning are mainly based on three foundational ideas. One of them is to constrain the parameter values to not deviate significantly from their  previously learned value by using some kind of regularization or a trade off between previous and new learned weights as in  \cite{progress,ewc, si, lp, imm}. A natural way to accomplish this is to train a model using online (at task-level) Bayesian inference, whereby the posterior of the parameters learned from task $t$ serve as the prior for task $t+1$ \cite{vcl, task_agnostic}. This informed prior helps in forward transfer, and also prevents catastrophic forgetting by penalizing large deviations from itself. In particular Variational Continual Learning \cite{vcl} (henceforth referred to as VCL ) achieves state of the art results by applying this simple idea to Bayesian neural networks. The second idea is to perform incremental model selection for every new task. For neural networks, this is done by evolving the structure as newer tasks are encountered \cite{cl_via_neural_prun,lrn2grow}. The third idea is to invoke a form of 'replay', whereby selected samples representative of previous tasks, are used to retrain the model after new tasks are learnt.

We present a novel Bayesian nonparametric approach to continual learning that seeks to incorporate the ability of structure learning into the simple yet effective framework of online Bayes. In particular, our approach models each hidden layer of the neural network using the Indian Buffet Process~\cite{ibp} prior, which enables us to grow the network dynamically as tasks arrive continually. We leverage the fact that any particular task uses a sparse subset of the connections of a neural network, and different related tasks share different (albeit possibly overlapping) subsets. Thus, in the setting of continual learning, it would be more effective if the neural network could accommodate changes in its connections to dynamically adapt to a newly arriving task. Moreover, in our model, we perform automatic model selection by letting each task select the number of nodes in each hidden layer. All this is done under the principled framework of variational Bayes and a nonparametric Bayesian modeling paradigm.

Another appealing aspect of our approach is that, unlike some of the recent state-of-the-art continual learning models~\cite{den,lrn2grow} that are specifically designed for supervised learning problems, our approach is applicable to both learning deep discriminative networks (supervised), where each task can be a Bayesian neural network~\cite{neal2012bayesian,bayesbybackprop}, as well as learning deep generative networks ({unsupervised)}, where each task can be a variational autoencoder~\cite{kingma2013auto}.
\vspace{-2em}
\section{Preliminaries}\label{sec:background}

\subsection{Bayesian Neural Networks and VAEs}
Bayesian neural networks~\cite{neal2012bayesian} are discriminative models where the goal is to model the relationship between inputs and outputs via a deep neural network with parameters $\bm{w}$. The network parameters are assumed to have a prior $p(\bm{w})$ and the goal is to infer the posterior given the observed data $\mathcal{D}$. Exact posterior inference is intractable in such models. One common approximate inference scheme is Bayes-by-Backprop \cite{bayesbybackprop} which uses a mean-field variational posterior $q(\bm w)$ over the weights. Note that $q(\bm w)$ is not restricted to be Gaussian. Reparameterized samples from this posterior are then used to approximate the lower bound via Monte Carlo sampling. Our goal in the continual learning setting is to learn such Bayesian neural networks  for a sequence of tasks by inferring the posterior $q_t(\bm{w})$ for each task $t$, without forgetting the information contained in the posteriors of previous tasks. %Using $S$ samples $\bm w^i \sim q(\bm w)$), we thus have

Variational autoencoders (VAE)~\cite{kingma2013auto} are generative models where the goal is to model a set of inputs $\{\bm{x}\}_{n=1}^N$ in terms of a stochastic latent variables $\{\bm{z}\}_{n=1}^N$. The mapping from each $\bm{z}_n$ to $\bm{x}_n$ is defined by a generator/decoder model (modeled by a deep neural network with parameters $\theta$) and the reverse mapping is defined by a recognition/encoder model (modeled by another deep neural network with parameters $\phi$). Inference in VAEs is done by maximizing the variational lower bound on the marginal likelihood. It is customary to do point estimation for decoder parameters $\theta$ and posterior inference for encoder parameters $\phi$. However, in the continual learning setting, it would be more desirable to infer the full posterior $q_t(\bm{w})$ for each task's encoder and decoder parameters $\bm{w} = \{\theta,\phi\}$, while not forgetting the information about the previous tasks as more and more tasks are observed. Our proposed continual learning framework address this aspect as well.

% \begin{align}\notag\label{bayes_by_backprop}
%     \mathcal L &= \mathbb{E}_{q(\bm{w})}\ln{\frac{p(\mathcal{D}, \bm{w})}{p(\bm{w})}} &\approx \frac{1}{S}\sum_{i=1}^{S}[\ln{p(\mathcal{D}|\bm{w}^i)} + p(\bm{w}^i) - q(\bm{w}^i))]
% \end{align}

%Bayesian neural networks~\cite{neal2012bayesian} were introduced with the objective of estimating the full posterior over the weights of a neural network in order to alleviate overfitting issues and provide predictive uncertainty estimates for the parameters of the neural network. Hence, for data $D$ and parameters $\bm{w}$, a prior $p(\bm{w})$ is imposed on the parameters, and we seek to learn $p(\bm{w}|D) = \frac{p(D|\bm{w})p(\bm{w})}{p(D)}$, where $p(D) = \int p(D|\bm{w})p(\bm{w})d\bm{w}$.
%The non-linearities of the neural network obliterate the possibilities of closed form updates; however, maximization of a variational lower bound via unbiased gradients provides a flexible means of doing approximate inference. Among other approaches, Bayes-by-Backprop \cite{bayesbybackprop} use a mean-field variational posterior $q(\bm w)$ over the weights. Note that $q(\bm w)$ is not restricted to be Gaussian. Reparameterized samples from this posterior are then used to approximate the lower bound via Monte Carlo sampling. Using $S$ samples $\bm w^i \sim q(\bm w)$) we thus have
% \begin{align}\notag\label{bayes_by_backprop}
%     \mathcal L &= \mathbb{E}_{q(\bm{w})}\ln{\frac{p(D, \bm{w})}{p(\bm{w})}} \\ &\approx \frac{1}{S}\sum_{i=1}^{S}[\ln{p(D|\bm{w}^i)} + p(\bm{w}^i) - q(\bm{w}^i))]
% \end{align}

\begin{figure}[!htbp]
  \centering
    \includegraphics[scale=0.14]{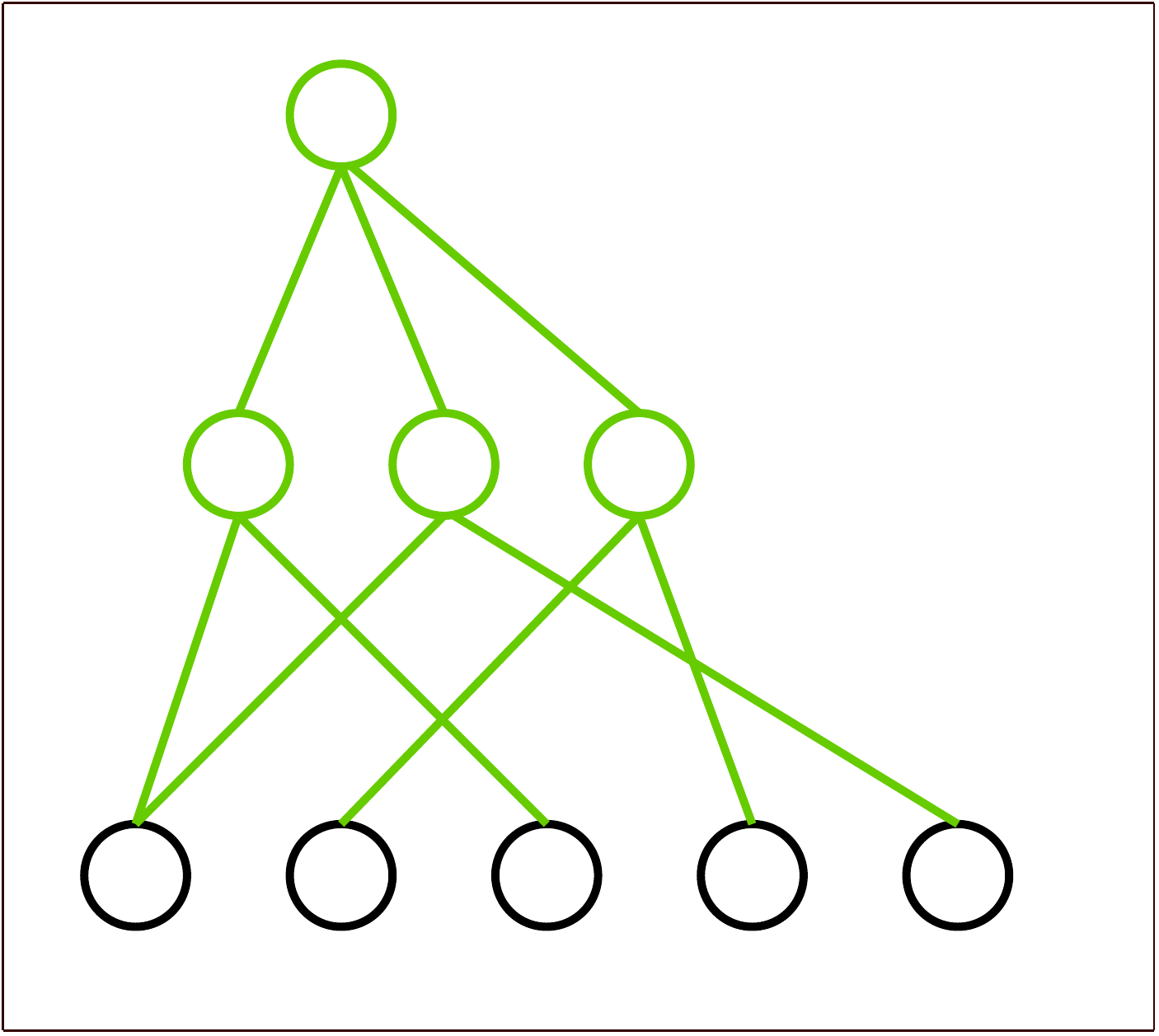}\hspace{1em}
    \includegraphics[scale=0.14]{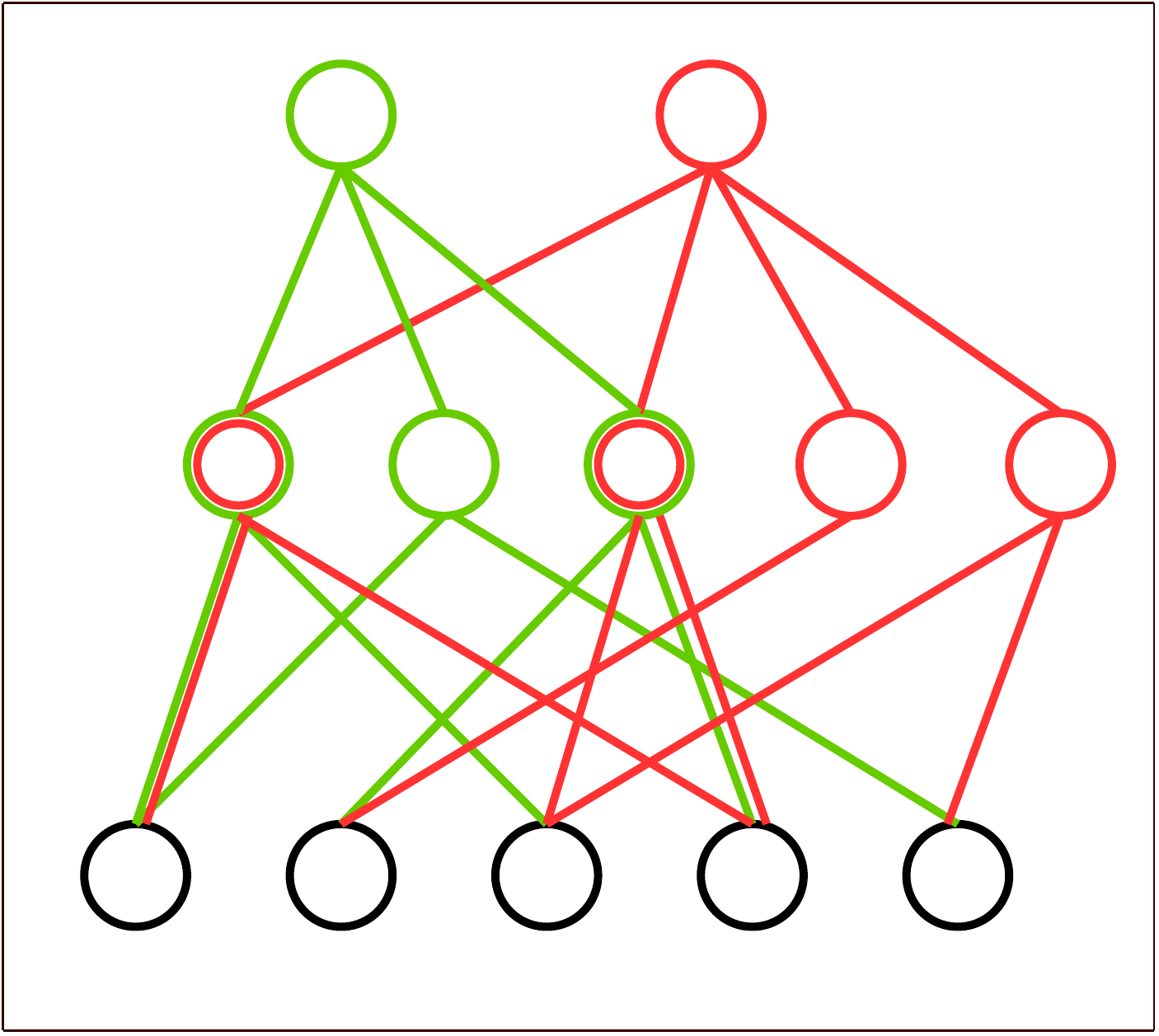}\hspace{1em}
    \includegraphics[scale=0.14]{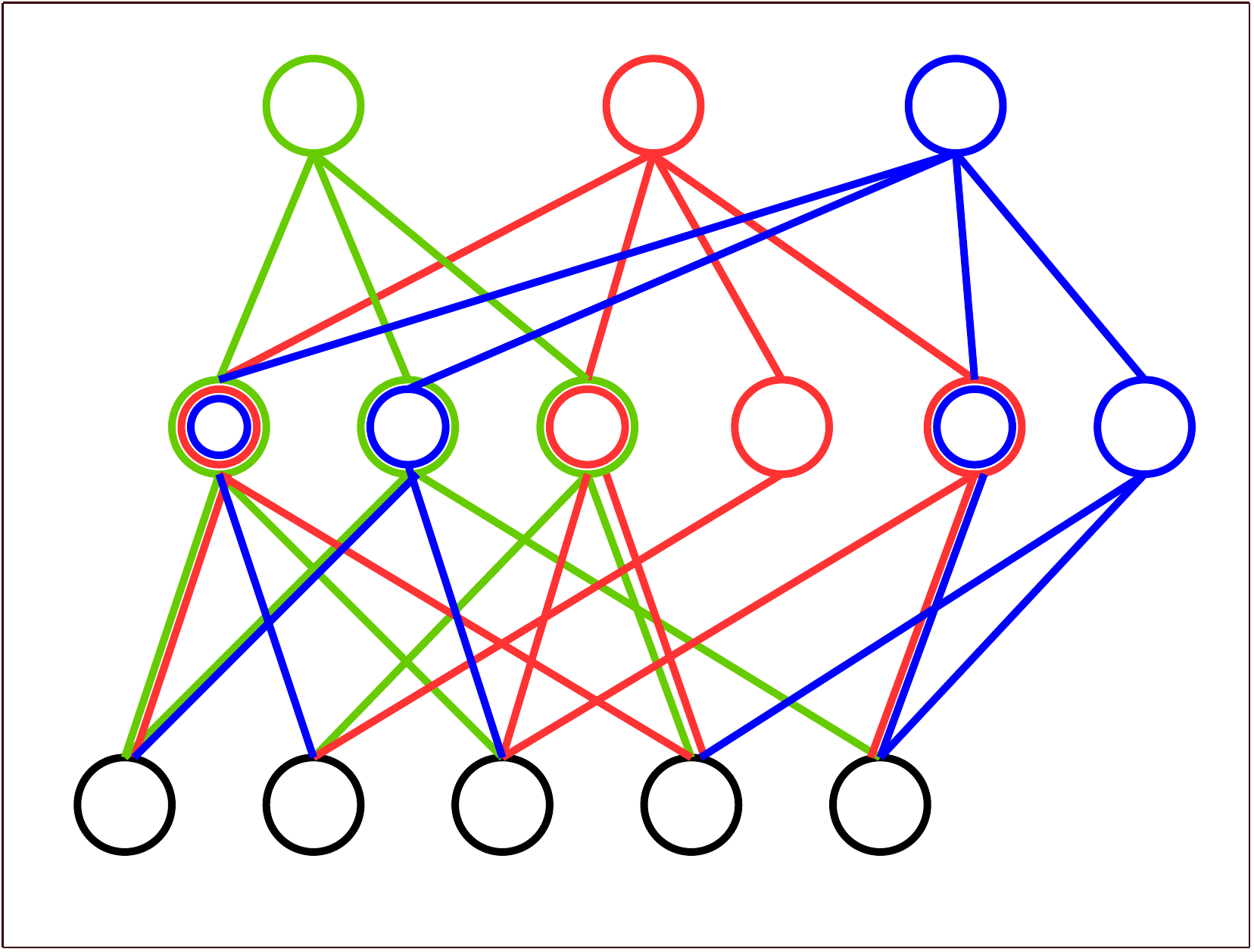}
  \caption{Evolution of network structure (with a single hidden layer) for 3 consecutive tasks. Weights/nodes used by task 1,2,3 are coloured green, red, blue respectively. Note that there can be overlapping of structure between tasks.
  }
     \label{fig:1}
\vspace{-1em}
\end{figure}

\subsection{Variational Continual Learning}

Variational Continual Learning (VCL)~\cite{vcl} is a recently proposed Bayesian approach to continual learning that combats catastrophic forgetting in deep neural networks by modeling the network parameters $\bm{w}$ in a Bayesian fashion and by setting $p_t(\bm w)=q_{t-1}(\bm w)$, that is, a task reuses the previous task's posterior as its prior. VCL solves the follow KL divergence minimization problem
\begin{equation}
    q_t(\bm{w}) = \arg\min_{q\in \mathcal{Q}}\text{KL}\left(q(\bm{w})||\frac{1}{Z_t}q_{t-1}(\bm{w})p(\mathcal{D}_t|\bm{w})\right)
\end{equation}

While offering a principled way that is applicable to both supervised (discriminative) and unsupervised (generative) learning settings, VCL assumes that the model structure/size is held fixed throughout, which can be limiting in continual learning where the number of tasks and their complexity is usually unknown beforehand. This necessitates adaptively inferring the model structure/size, that can potentially adapt/grow with each incoming task. Another limitation of VCL is that the unsupervised version, based on performing CL on VAEs, only does so for the decoder model's parameters (shared by all tasks). It uses completely task-specific encoders and, consequently, is unable to transfer information across tasks in the encoder model.  Our proposed Bayesian framework addresses both these limitations in a principled manner.

\section{A Nonparametric Bayesian Approach to Continual Learning}
We present a nonparametric Bayesian model for continual learning that can potentially grow and adapt its structure as more and more tasks are observed. Our model also extends seamlessly for unsupervised learning as well. For brevity of exposition, in this section, we mainly focus on the supervised setting. We briefly discuss the unsupervised extension (based on VAEs) in Sec.~\ref{sec:unsup} and provide further details of the unsupervised extension in the Supplementary Material.

\begin{figure}[t!]
  \centering
    \includegraphics[scale=0.30,trim={0em 0em 0em 0em},clip]{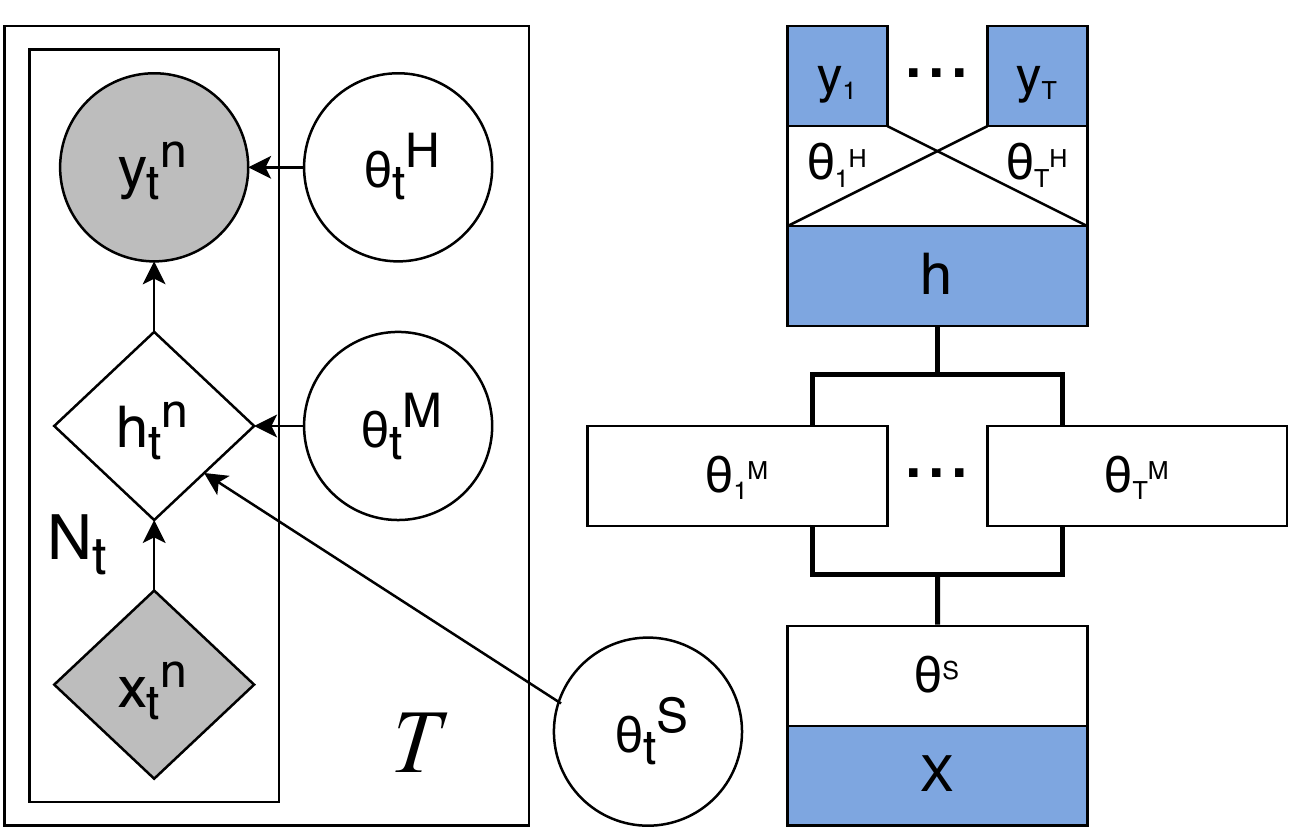}
    \includegraphics[scale=0.30,trim={0em 0em 0em 0em},clip]{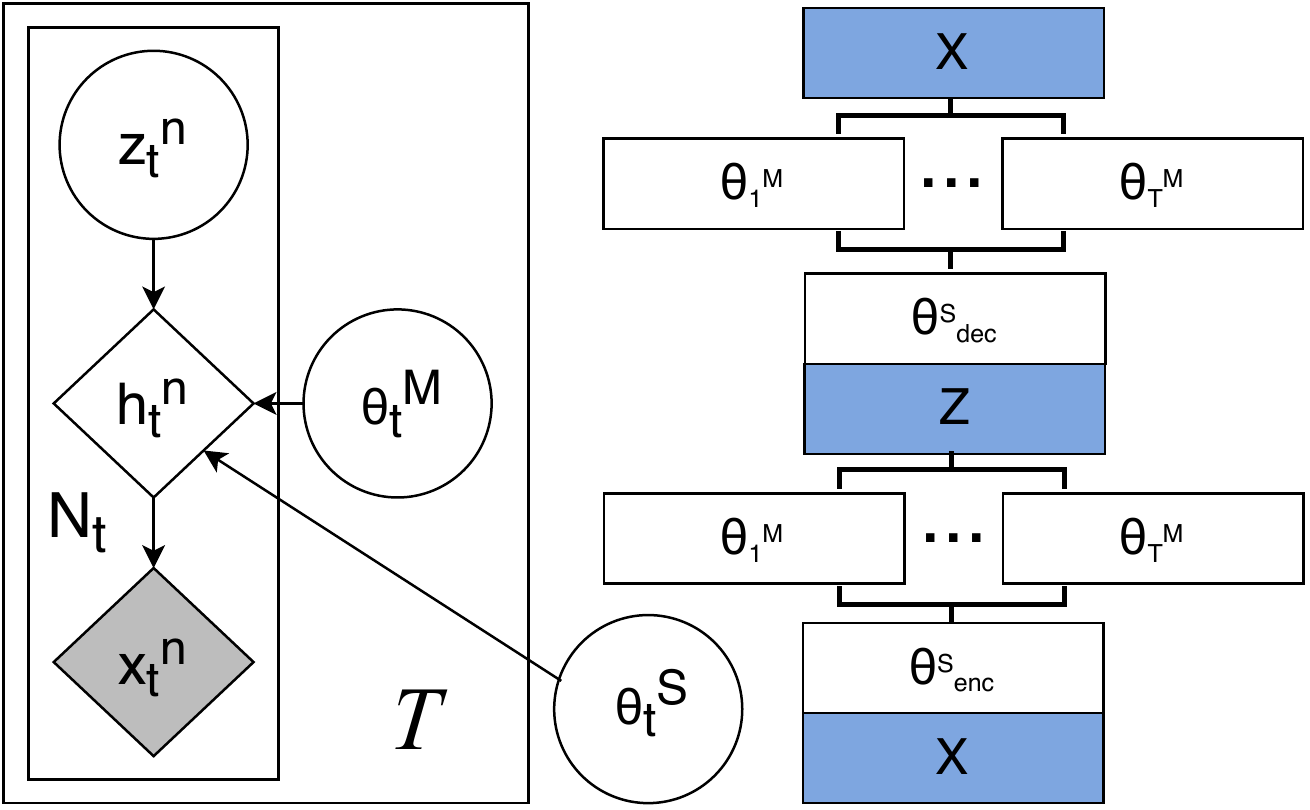}
    \begin{minipage}{0.45\linewidth}
        \centering
        (a)
    \end{minipage}
    \begin{minipage}{0.45\linewidth}
        \centering
        (b)
    \end{minipage}
    \vspace{-1em}
  \caption{Schematics representing our models. In both (a) discriminative model and (b) generative model, $\theta_S$ are parameters shared across all task, $\theta^M$ are the task specific mask parameters, and $\theta^H$ are last layer seperate head parameters. In our exposition, we collectively denote these parameters by $\bm W = \bm B \odot \bm V$ with the masks being $\bm B$ and other parameters being $\bm V$. 
  }
     \label{fig:schema}
\vspace{-1em}
\end{figure}

Our model is based on using a basic primitive that models each hidden layer using a nonparametric Bayesian prior (Fig.~\ref{fig:1} shows an illustration and Fig.~\ref{fig:schema} shows a schematic diagram). These hidden layers can be used in Bayesian neural networks to model the feedforward connections, or in VAEs for the decoder and encoder models. Assuming a single hidden layer for simplicity, the first task allocates as many hidden layer nodes as necessary, and learns the posterior over weights for a subset of the edges incident on each node. Each subsequent task reuses some of the edges learnt by the previous task and uses the posterior over the weights learnt by the previous task as the prior. Additionally, it may allocate new nodes and learn the posterior over some of their incident edges. It thus learns the posterior over ($i$) a subset of the weights used by the previous task, ($ii$) a subset of the weights (incident on previously existing nodes) that were \textit{not} used by the previous task, ($iii$) weights of a subset of the connections incident on the  nodes newly allocated by itself. While making predictions, a task uses only the nodes/weights it has learnt. More slack for later tasks in terms of model size (allowing it to create new nodes) indirectly lets the task learn better without deviating too much from the prior, which in this case is the posterior of the previous tasks. This reduces chances of catastrophic forgetting~\cite{ewc}.

\subsection{Generative story}
Omitting the task id $t$ for brevity, consider modeling this task using a neural network having $L$  hidden layers. We model the weights in layer $l$ as $\bm W^l = \bm{B}^l \odot \bm{V}^l$, a point-wise multiplication of a real-valued matrix $\bm V^l$ (with a Gaussian prior $\mathcal{N}(0,\sigma_0^2)$ on each entry) and a binary matrix $\bm B^l$. This ensures sparse connection weights between the layers. Moreover, we model $\bm B^l \sim \text{IBP}(\alpha)$ using the Indian Buffet Process (IBP)~\cite{ibp} prior, where the hyperparameter $\alpha$ controls the number of nonzero columns in $B$ and its sparsity. The IBP prior thus enables learning the size of $\bm B^l$ (and consequently of $\bm V^l$) from data. As a result, the number of node in the hidden layer is learned adaptively from data. The output layer weights are denoted as $\bm W_{out}$ with each weight having a Gaussian prior $\mathcal{N}(0,\sigma_0^2)$. The outputs are then assumed to be generated as %$\bm{B}^l \sim \text{IBP}(\alpha)$, $\bm{V}_{d,k}^l \sim \mathcal N(0, \sigma_0^2)$, $\bm{W}^l = \bm{B}^l \odot \bm{V}^l$, $\bm{W}_{d,k}^{out} \sim \mathcal N(0, \sigma_0^2)$, and $\bm{y}_n \sim \text{Lik}(\bm{W}_{out}\bm{\phi}_{NN}(\bm{x}_n))$,
\begin{align}
    % \bm{B}^l &\sim \text{IBP}(\alpha)\\
    % \bm{V}_{d,k}^l &\sim \mathcal N(0, \sigma_0^2)\\
    % \bm{W}^l &= \bm{B}^l \odot \bm{V}^l\\
    % \bm{W}_{d,k}^{out} &\sim \mathcal N(0, \sigma_0^2)\\
    \bm{y}_n &\sim \text{Lik}(\bm{W}_{out}\bm{\phi}_{NN}(\bm{x}_n)), n=1,\ldots,N
\end{align}
Here $\bm \phi_{NN}$ is the function computed (using parameter samples) up to the last hidden layer of the network thus formed, and $\operatorname{Lik}$ denotes the likelihood model for the outputs.

Similar priors on the network weights have been used in other recent works to learn sparse deep neural networks~\cite{panousis2019nonparametric,xu2019variational}. However, these works assume a single task to be learned. In contrast, our focus here is to leverage such priors in the continual learning setting where we need to learn a sequence of tasks, while avoiding the problem of catastrophic forgetting. %Connection with the above model can also be traced to \cite{mtlinfinite} which uses an IBP prior on the shared latent basis of the weight vectors in a linear multitask learning model, allowing the size of the basis to be learned from data.

Henceforth, we further suppress the superscript denoting layer number from the notation for simplicity; the discussion will hold identically for all hidden layers.

When adapting to a new task in our continual learning setting, the posterior of $\bm V$ learnt from previous tasks is used as the prior. A new $\bm B$ is learnt afresh, to ensure that a task only learns the subset of weights relevant to it. As described before, to adaptively infer the number of nodes in each hidden layer, we use the IBP prior \cite{ibp}, whose truncated stick-breaking process \cite{doshivelezibp} construction for each entry of $B$ is as follows
\begin{align}\label{stick_breaking}
    \nu_{k} &\sim \text{Beta}(\alpha,1), \quad \pi_k = \prod_{i=1}^{k}\nu_i\\
    B_{d,k} &\sim \text{Bernoulli}(\pi_k)
\end{align}
% \begin{align}\label{stick_breaking}
%     \nu_{k} &\sim \text{Beta}(\alpha,1), \quad \pi_k = \prod_{i=1}^{k}\nu_i, \quad
%     B_{d,k} \sim \text{Bern}(\pi_k)
% \end{align}
for $d \in 1,...,D$, where $D$ denotes the number of input nodes for this hidden layer, and $k \in 1, 2,..., K$, where $K$ is the truncation level and $\alpha$ controls the effective value of $K$, i.e., the number of active hidden nodes. Note that the prior probability $\pi_k$ of weights incident on hidden node $k$ being nonzero decreases monotonically with $k$, until, say, $K$ nodes, after which no further nodes have any incoming edges with nonzero weights from the previous layer, which amounts to them being turned off from the structure. Moreover, due to the cumulative product based construction of the $\pi_k$'s, an implicit ordering is imposed on the nodes being used. This ordering is preserved across tasks, and allocation of nodes to a task follows this, facilitating reuse of weights.

The truncated stick-breaking approximation is a practically plausible and intuitive solution for continual learning, since a fundamental tenet of continual learning is that the model complexity should not increase in an unbounded manner as more and more tasks are encountered. Suppose  we fix a budget on the maximum allowed size of the network (say, the number of hidden nodes allowed in each layer) after it has seen, say, $T$ tasks. This exactly corresponds to the truncation level for each layer. Then for each task, nodes are allocated conservatively from this total budget, in a fixed order, conveniently controlled by the $\alpha$ hyper parameter. In Sec.~\ref{sec:dynexp}, we also discuss a dynamic expansion scheme that avoids specifying a truncation level.

\subsection{Inference}
Exact inference is intractable in this model due to non-conjugacy. Therefore, we resort to variational inference~\cite{blei2017variational}. We employ structured mean-field approximation~\cite{svi}, which performs better than normally used mean-field approximation, as the former captures the
dependencies in the approximate posterior distributions of $\bm{B}$ and $\bm{\nu}$. In particular, we use %. We use a variational posterior family of the following form 
% -------------------------------------------------
% WRITE ABOUT THE STRUCTURED USE OF INFERENCE
% -------------------------------------------------
% \vspace{-0.5mm}
\begin{align}
q(\bm V, \bm B, \bm v) = q(\bm V) q(\bm B | v) q(\bm v)
\end{align}
where, $q(\bm V) = \prod_{d=1}^D\prod_{k=1}^K \mathcal{N}(V_{d,k}|\mu_{d,k}, \sigma_{d,k}^2)$ is a mean field Gaussian approximation for the weights. Corresponding to the Beta-Bernoulli hierarchy of (\ref{stick_breaking}), we use the conditionally factorized variational posterior family , that is, $q(\bm B | v) = \prod_{d=1}^D \prod_{k=1}^K \operatorname{Bern}(B_{d,k}| \theta_{d,k})$, where $\theta_{d,k} = \sigma(\rho_{d,k} + logit(\pi_k))$ and $q(\bm v) = \prod_{k=1}^K \operatorname{Beta}(v_k | \nu_{k,1}, \nu_{k,2})$

Thus we have $\Theta = \{  \nu_{k,1}, \nu_{k,2}, \{\mu_{d,k}, \sigma_{d,k},\rho_{d,k}\}_{d=1}^D \}_{k=1}^K$ as the complete set of learnable variational parameters.

% \\
Each column of $\bm B$ represents the binary mask for the weights incident to a particular node. Note that although these binary variables (in a single column of $\bm B$) share a common prior, the posterior for each of these variables is different, thereby allowing a task to selectively choose a subset of the weights leading to an activation, with the common prior controlling the degree of sparsity.

Bayes-by-backprop~\cite{bayesbybackprop} is a common choice for performing variational inference in this context. The Evidence Lower Bound (ELBO) can be expressed via the data-dependent likelihood and data-independent KL terms
\begin{align} \label{elbo}\notag
    \mathcal{L} =  &\ \mathbb{E}_{q(\bm V, \bm B, \bm v)}[\ln p(\bm Y|\bm V, \bm B, \bm v)]\\ &- KL(q(\bm V, \bm B, \bm v) || p(\bm V, \bm B, \bm v))
\end{align}
Using the factorization of the joint prior $p(\bm V, \bm B, \bm v)=p(\bm V) p(\bm B| \bm v) p(\bm v)$ and the mean-field factorization of the posterior, the KL divergence term of (\ref{elbo}) decomposes as
\begin{align} \label{KL_term}\notag
 KL[q(\bm V)||p(\bm V)] &+ \mathbb{E}_{q(\bm v)}[KL[q(\bm B | \bm v)||p(\bm B|\bm v)]]\\ &+ KL[q(\bm v)|| p(\bm v)]
\end{align}

All the KL divergence terms in the above expression have closed form expressions; hence using them directly rather than estimating them from Monte Carlo samples alleviates the approximation error as well as the computational overhead due to sampling, to some extent.
The expectation terms are optimized by unbiased gradients from the respective posteriors. Using Bayes-by-backprop, we thus have
\begin{align}\notag
\mathcal{L} = &\frac{1}{S}\sum_{i=1}^S [f_{l}(\bm V^i, \bm B^i, \bm v^i) - KL[q(\bm B|\bm v^i)||p(\bm B|\bm v^i)]] \\&- KL[q(\bm V)||p(\bm V)] - KL[q(\bm v)||p(\bm v)]
\label{eq:9}
\end{align}

The log-likelihood term is decomposed as 
\begin{align}
    f_l(\bm V, \bm B, \bm v) &=\log \operatorname{Lik}(\bm Y|\bm V, \bm B, \bm v)
    \\&=\log \operatorname{Lik}(\bm Y | \bm W_{out} \bm \phi_{NN}(\bm{X; V, B, v}))\notag
\end{align}

where $(\bm X, \bm Y)$ is the training data. For regression, $\operatorname{Lik}$ can be Gaussian with some noise variance, while for classification it can be Bernoulli with a probit or logistic link.

\paragraph{Sampling details} We obtain unbiased reparameterized gradients for all the parameters of the variational posterior distributions. For the Bernoulli distributed variables, we employ the Gumbel-softmax trick \cite{gumbsoft}, also known as CONCRETE \cite{concrete}. For Beta distributed $v$'s, the Kumaraswamy Reparameterization Gradient technique \cite{kumara} is used. For the real-valued weights, the standard location-scale trick of Gaussians is used. The Supplementary Material contains detailed equations.

\subsection{Unsupervised Continual Learning}
\label{sec:unsup}
Our discussion thus far has primarily focused on continual learning where each task is a supervised learning problem. Our framework however readily extends to unsupervised continual learning~\cite{vcl,smith2019unsupervised,rao2019continual} where we assume that each task involves learning a deep generative model, commonly a VAE~\cite{vcl,smith2019unsupervised,rao2019continual}. In this case, each input observation $\bm{x}_n$ has an associated latent variable $\bm{z}_n$. Collectively denoting all inputs as $\bm X$ and all latent variables as $\bm Z$, we can define an ELBO similar to Eq.~\ref{elbo} as follows :
\begin{align} \label{elbo1}\notag
    \mathcal{L} =  &\ \mathbb{E}_{q(\bm Z, \bm V, \bm B, \bm v)}[\ln p(\bm X|\bm Z, \bm V, \bm B, \bm v)]\\ &- KL(q(\bm Z, \bm V, \bm B, \bm v) || p(\bm Z, \bm V, \bm B, \bm v))
\end{align}
Note that, unlike the supervised case, the above ELBO also involves an expectation over $\bm Z$.
Similar to Eq.~\ref{eq:9} this can be approximated using Monte Carlo samples, where each ${\bm z}_n$ is sampled from the amortized posterior $q({\bm z}_n|\bm V,\bm B,\bm v,{\bm x}_n)$. In addition to learning the model size adaptively, as shown in the schematic diagram (Fig.~\ref{fig:schema} (b)), our model learns shared weights and task-specific masks for the encoder and decoder models. In contrast, VCL~\cite{vcl} uses fixed-sized model and entirely task-specific encoders (and of pre-defined sizes), which prevents knowledge transfer across the different encoders. 

\section{Other Key Considerations}
\label{sec:keycon}
In continual learning setting where the goal is to learn a sequence of tasks, a few other aspects deserve additional consideration. In this section, we discuss how we incorporate them in the context of our proposed model.
%The model described in the previous section is almost completely specified for a single task. As such, it is by itself interesting as a Bayesian neural net model that adapts its architecture, driven by the task at hand. However, as evidenced by recent works (see Section \ref{related}), it is the  setting of continual learning that can benefit tremendously from such ability to learn model structure. In this section, we therefore focus on applying the model in this setting.

\subsection{Masked Priors}
Using previous task's posterior as the prior for current task~\cite{vcl} may be problematic in some cases. 
%While learning the tasks sequentially, using some assumed prior like zero mean Gaussian in calculation of KL divergence increases learning complexity. 
For example, the partially learned parameters that do not contribute to previous task may not be useful to be part of the prior for the current task. In fact, instead of forward transfer, using them as prior for next task might even promote catastrophic forgetting. To overcome this issue, we mask the new prior for next task with initial prior as
\begin{align}\label{mask}
     p_t(\bm V_{d,k}) &= B_{d,k}^oq_{t-1}(\bm V_{d,k}) + (1-B_{d,k}^o)p_0(\bm V_{d,k})
\end{align}
where ${\bm B}^o$ is the overall combined mask from all previously learnt tasks i.e., (${\bm B}^1 \cup {\bm B}^2...\cup {\bm B}^{t-1}$), $q_{t-1}, p_t$ are the previous posterior and current prior, respectively, and $p_0$ is the prior used for first task. This makes sense as the partially trained weights will cause undesirable regularization for next task as it does not help retaining the previous tasks performance. Standard choices of initial prior $p_0$, such as a zero mean normal distribution or uniform distribution with this masking, further reduces the catastrophic forgetting by promoting the use of new weights or weights with higher variance in previously learned tasks.
\vspace{-1em}
\subsection{Segregating the head} \label{head}
It has been shown in prior work on supervised continual learning \cite{task_agnostic} that using separate last layers (commonly referred to as ``heads'') for different tasks dramatically improves performance in continual learning. 
%The individual tasks in continual learning are typically multiclass classification problems, with each task classifying between 2 classes (see section \ref{exp}). 
Therefore, in the supervised setting, we use a generalized linear model that uses the embeddings from the last hidden layer, with the parameters up to the last layer involved in transfer and adaptation. 

% \begin{algorithm}[tb]
% \caption{Nonparametric Bayesian CL}\label{algo}
% \begin{algorithmic}
% \STATE {\bfseries Input:}{Initial Prior $p_0(\Theta)$} ; $p_{new} \leftarrow p_0(\Theta)$
% \STATE Initialize the network parameters and coresets. 
%     \FOR{$i=1$ {\bfseries to} $T$}
%     \STATE Observe current task data $D_t$\;
%     \STATE Update coresets \ref{coreset}\;
%     \STATE \textbf{Masked Training}\;
    
%     \begin{ALC@g}
%     \STATE $\mathcal L_t \leftarrow$ ELBO (eq. \ref{elbo}) with prior $p_{new}$\;
%     \STATE $\Theta_t \leftarrow \arg \min \mathcal L_t$\;
%     \end{ALC@g}
%     \STATE \textbf{Selective Finetuning}\;
%     \begin{ALC@g}
%     \STATE Fix the IBP parameters and learned mask\;
%     \STATE $\Theta_t \leftarrow \arg \min \mathcal L_t$\;
%     \end{ALC@g}
%     \STATE $p_{new} \leftarrow q_t(\Theta)$\;
%     \STATE $p_{new} \leftarrow$ Mask($p_{new}$) using eq \ref{mask}\;
%     \STATE Perform prediction for given test set.
%     \ENDFOR
% \end{algorithmic}
% \end{algorithm}
\vspace{-1em}
\subsection{Prediction-driven training with coresets} \label{coreset}
Proposed in~\cite{vcl} as a method for cleverly sidestepping the issue of catastrophic forgetting, the coreset comprises representative training data samples from all tasks. Let $M^{(t-1)}$ denote the posterior state of the model before learning task $t$. With the $t$-th task's arrival having data $D_t$, a coreset $C_t$ is created comprising choicest examples from tasks $1\dots t$. Using data $D_t \setminus C_t$ and having prior $M^{(t-1)}$, new model posterior $M^t$ is learnt. For predictive purposes at this stage (the test data comes from tasks $1\dots t$), a new posterior $M^t_{pred}$ is learnt with $M^t$ as prior and with data $C_t$. Note that $M^t_{pred}$ is used only for predictions at this stage, and does not have any role in  the subsequent learning of, say, $M^{(t+1)}$. Such a predictive model is learnt after every new task, and discarded thereafter. Intuitively it makes sense as some new learnt weights for future tasks can help the older task to perform better (backward transfer) at testing time. For more details, please refer to Appendix C.

During the coreset-based training phase after task $t$, we only update the weights for the tasks $1, \dots ,t-1$ with (and using) the binary mask fixed at its previously learned value, i.e., a task refines only its own subset of weights.
\vspace{-1em}
\subsection{The IBP hyperparameter $\alpha$}
Although we found using a sufficiently large value of $\alpha$ without tuning to perform reasonably, we also considered using a schedule with $\alpha$ increasing gradually, and the possibility of learning $\alpha$. We discuss the further details in the Supplementary Material. %tasks several considerations can be made regarding its choice.

%\paragraph{Scheduling across tasks} Intuitively, $\alpha$ should be incremented for every new task according to some schedule. Information about task relatedness can play a decisive role in formulating the schedule. Smaller increments of $\alpha$ discourage creation of new nodes and encourage more sharing of already existing connections across tasks. 

%\paragraph{Learning $\alpha$} Although not investigated in this work, one viable alternative to choosing $\alpha$ by cross-validation could be to learn it. This can be easily accommodated into our variational framework by imposing a gamma prior on $\alpha$ and using a suitably parameterized gamma variational posterior. The only difference in the objective would be in the KL terms: the KL divergence of $\bm v$ will then also have to estimated by Monte Carlo approximation (because of dependency on $\alpha$ in the prior). Also, since gamma distribution does not have an analytic closed form KL divergence, the Weibull distribution can be a suitable alternative~\cite{zhang2018whai}.

\subsection{Other Practical issues}
\paragraph{Space complexity} The proposed scheme entails storing a binary matrix for each layer of each task which results into 1 bit per weight parameter, which is not very prohibitive and can be efficiently stored/compressed in sparse matrices. Moreover, the initial tasks make use of only a limited number of the first few columns of the IBP matrix, and hence does not pose any significant overhead.

\paragraph{Adjusting bias terms} The IBP selection acts on the weight matrix only. For the hidden nodes \textit{not} selected in a task, their corresponding biases need to be removed as well. In principle, the bias vector for a hidden layer should be multiplied by a binary vector $\bm u$, with $u_i=\mathbb{I}[\exists d: B_{d,i}=1]$. In practice, we simply scale each bias component by the maximum reparameterized Bernoulli value in that column.

\paragraph{Selective Finetuning}
While training with reparameterization (Gumbel-softmax), the learnt masks are close to binary but not completely binary which affects task performance a bit. So we fine-tune the network with fixed structure (i.e Beta-Bernoulli distributions parameters are fixed) after it has been learned, to restore the accuracy of the task. %A summarized version of 
%Algorithm \ref{algo} summarizes our models training procedure. The method for update of coresets that we used are similar to as it was proposed in \cite{vcl}.

\subsection{Dynamic Expansion}
\label{sec:dynexp}
Although our inference scheme uses a truncation-based approach for the IBP posterior, it is possible to do inference in a truncation-free manner. One possibility is to greedily increase layer width until performance saturates. However we empirically found that this leads to a bad optima. We can leverage the fact that, given a sufficiently large number of columns, the last columns of the IBP matrix tends to be all zeros. So we increase the number of hidden nodes after every iteration to keep the number of such empty columns equal to a constant value $\mathcal T^{l}$ in following manner. 
\begin{equation}
C_j^l = C_{j+1}^{l}\prod_{i}^{D^l}\mathbb{I}(B_{ij}^{l}=0), \quad G^{l} = \mathcal T^{l} - \sum_{j=1}^{K^l} C_{j}^{l}
\end{equation}
where $l$ represents current layer index, $B^l$ is the sampled IBP mask for current task, $C_j^l$ indicates if all columns from $j^{th}$ column onward are empty. $G^l$ is the number of hidden units to expand in the current network layer.

\section{Related Work} \label{related}
One of the key challenges in continual learning (henceforth referred to as CL) is to prevent catastrophic forgetting, typically addressed through regularization of the parameter updates, preventing them from drastically changing from the value learnt from the previous task(s). Notable methods based on this strategy include EwC \cite{ewc}, SI \cite{si}, Laplace approximation \cite{lp}, etc. Superceding these methods is the Bayesian approach, a natural remedy of catastrophic forgetting in that, for any task, the posterior of the model learnt from the previous task serves as the prior for the current task, which is the canonical online Bayes. This approach is utilized in recent works like VCL \cite{vcl} and task agnostic variational Bayes \cite{task_agnostic} for learning Bayesian neural networks in the CL setting. Our work is most similar in spirit to and builds upon this body of work.

Another key aspect in CL methods is \emph{replay}, where some samples from previous tasks (selected randomly or by some heuristic) are used to fine-tune the model after learning a new task (thus refreshing its memory in some sense and avoiding catastrophic forgetting). Some of the works using this idea include \cite{gem}, which solves a constrained optimization problem at each task, the constraint being that the loss should decrease monotonically on a  heuristically selected replay buffer; \cite{modadapt}, which uses a partially shared parameter space for inter-task transfer and  \textit{generates} the replay samples through a data-generative module; and \cite{func_reg_gp}, which learns a Gaussian process for each task, with a shared mean function in the form a feedforward neural network, the replay buffer being the set of inducing points typically used to speed up GP inference. For VCL \cite{vcl} and our work, the coreset (section \ref{coreset}) serves as a replay buffer; but we emphasize that it is not the primary mechanism to overcome catastrophic forgetting in these cases, but rather an additional mechanism to preventing it.

Recent work in CL has investigated allowing the structure of the model to dynamically change with newly arriving tasks. Among these, strong evidence in support of our assumptions can be found in \cite{cl_via_neural_prun}, which also learns different sparse subsets of the weights of each layer of the network for different tasks. The sparsity is enforced by a combination of weighted $L_1$ regularization and threshold-based pruning. There are also methods that do not learn subset of weights to be used but rather learn the subset of hidden layer node outputs to be used for each task; such a strategy is adopted by either using Evolutionary Algorithms to select the node subsets \cite{pathnet} or by training the network with task embedding based attention masks \cite{hat}. One recent approach \cite{claw}, instead of using binary masks, tries to adapts weights at different scales for different tasks; it is also designed only for discriminative tasks.

Among other related work, \cite{lrn2grow,den,rcl} either reuse the parameters of a layer, dynamically grows the size of the hidden layer, or adapt them, \textit{or} spawn a new set of parameters (the model complexity being bounded through regularization terms or reward based reinforcements). Most of these approaches however tend to be rather expensive and rely on techniques, such as neural architecture search. In another recent work (simultaneous development with our work), \cite{kessler2019indian} did a preliminary investigation on using IBP for continual learning. They however use IBP on hidden layer activations instead of weights (which they mention is worth considering), do not consider issues such as the ones we discussed in Sec.~\ref{sec:keycon}, and only considered the supervised setting. % nonparametric Bayesian modeling and offers a principled and efficient way of learning a dynamically growing network. Moreover, our approach works for both supervised as well as unsupervised learning.  

%We would like to note here that usage of IBP prior for continual learning has been recently also explored, albeit briefly, in~\cite{kessler2019indian} (which is a simultaneous development). Moreover, our construction has several differences from theirs: (1) They use IBP on the hidden layer activations whereas we model the weights using IBP; (2) Their construction is limited to supevised learning while ours works for both supervised and unsupervised learning; and (3) Their results/evaluation study was preliminary whereas we perform an extensive evaluation.

%In comparison, our model already has a fixed budget in advance, which is motivated by the desideratum of CL that the model complexity should not increase in an unbounded manner as more and more tasks are encountered.
\vspace{-1em}
\section{Experiments} \label{exp}

We perform experiments on both supervised and unsupervised CL and compare our method with relevant state-of-the-art methods. In addition to the quantitative (accuracy/log-likelihood comparisons) and qualitative (generation) results, we also examine the network structures learned by our model. Some of the details (e.g., experimental settings) have been moved to the Supplementary Material\footnote{The code for our model can be found at this link: \url{https://github.com/scakc/NPBCL}}. 

\subsection{Supervised Continual Learning}
We first evaluate our model on standard supervised CL benchmarks. We experiment with different existing approaches such as, Pure Rehearsal \cite{rehearsal}, EwC \cite{ewc}, IMM \cite{imm}, DEN \cite{den}, RCL \cite{rcl}, and ``Na\"ive'' which learns a shared model for all the tasks.

We perform our evaluations on three supervised CL benchmark datasets: SplitMNIST, Split notMNIST(small), Permuted MNIST and fashion MNIST. For Split MNIST, the tasks consist of 5 binary classification problems, the splits being 0/1, 2/3, 4/5, 6/7, 8/9 digits. Split notMNIST consists of 5 binary classification splits as A/B, C/D, E/F, G/H, I/J and, similarily, fashionMNIST consists of 5 binary classification splits as T-shirt/Trouser, Pullover/Dress, Coat/Sandals, Shirt/Sneaker, and Bag/Ankle boots. For Permuted MNIST, each task is a multiclass classification problem. However, for each task, a fixed random permutation is applied to the pixels of the images of all classes. We generated 5 such tasks for our experiments. The heads are separate for different tasks (Sec.~\ref{head}).

\subsubsection{Performance evaluation}
Suppose we have $T$ tasks arriving sequentially. To gauge the effectiveness of our model towards preventing catastrophic forgetting, we report ($i$) the test accuracy of task $t$ after learning each of the subsequent tasks ($t, t+1, t+2, \dots T$); and ($ii$) the average test accuracy over all previous tasks $1,2,\dots t$ after learning each task $t$.

\begin{figure}[h!]
\centering
%\vspace{-1.0em}
\includegraphics[width=1.1\linewidth,trim={1em 0mm 0em 4em},clip]{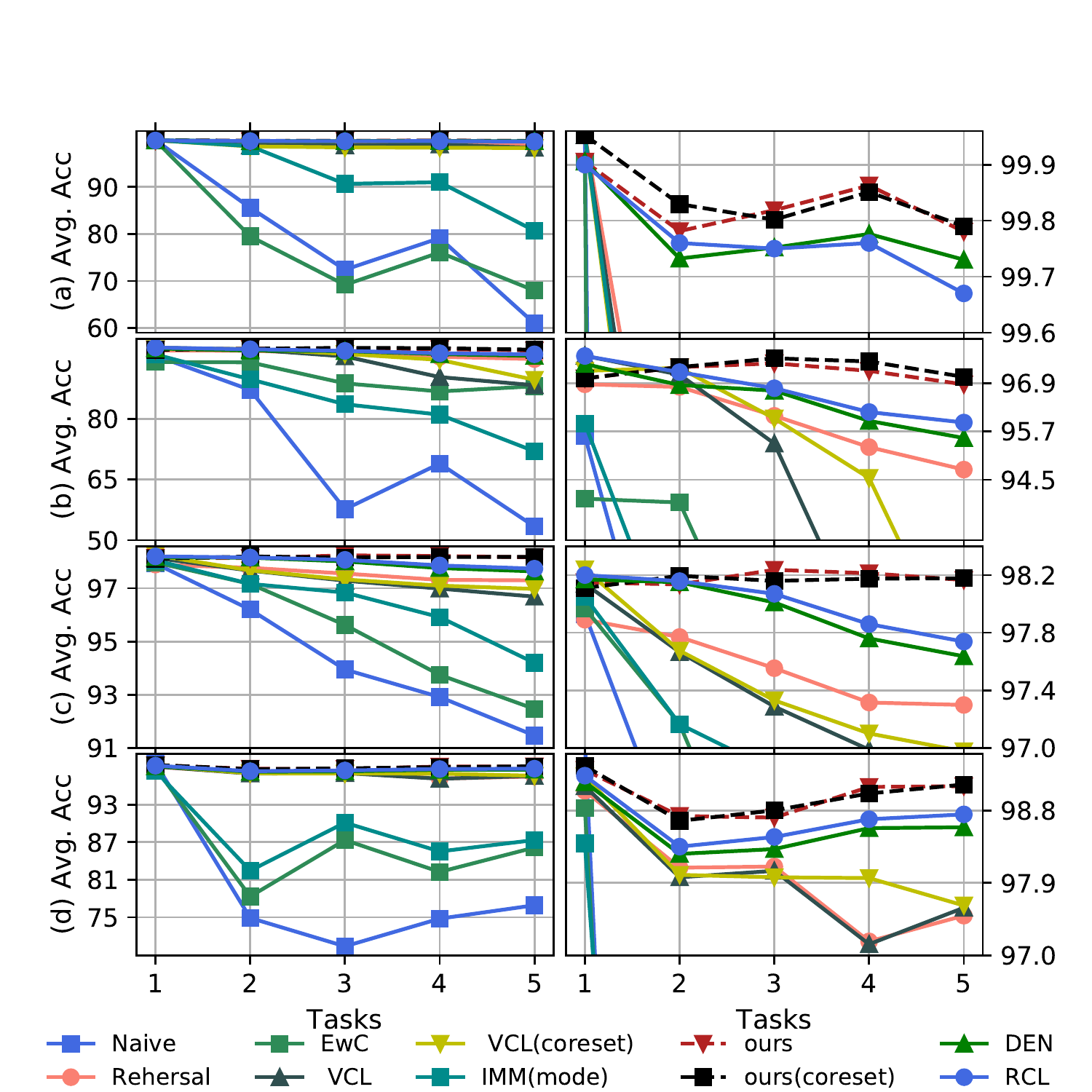}
\vspace{-1.5em}
\caption{\small{Mean test accuracies of tasks seen so far as new tasks are observed, for (a)splitMNIST, (b)notMNIST , (c)Permuted MNIST and (d) fashionMNIST, here (\textbf{right}) side plots are scaled version of (\textbf{left}) ones for better insight \vspace{-1.5em}}
}
\label{fig:myfig}
\end{figure}

\begin{figure}[!htbp]
    \centering
    \includegraphics[width=0.8\linewidth,trim={0em 0em 0em 0em},clip]{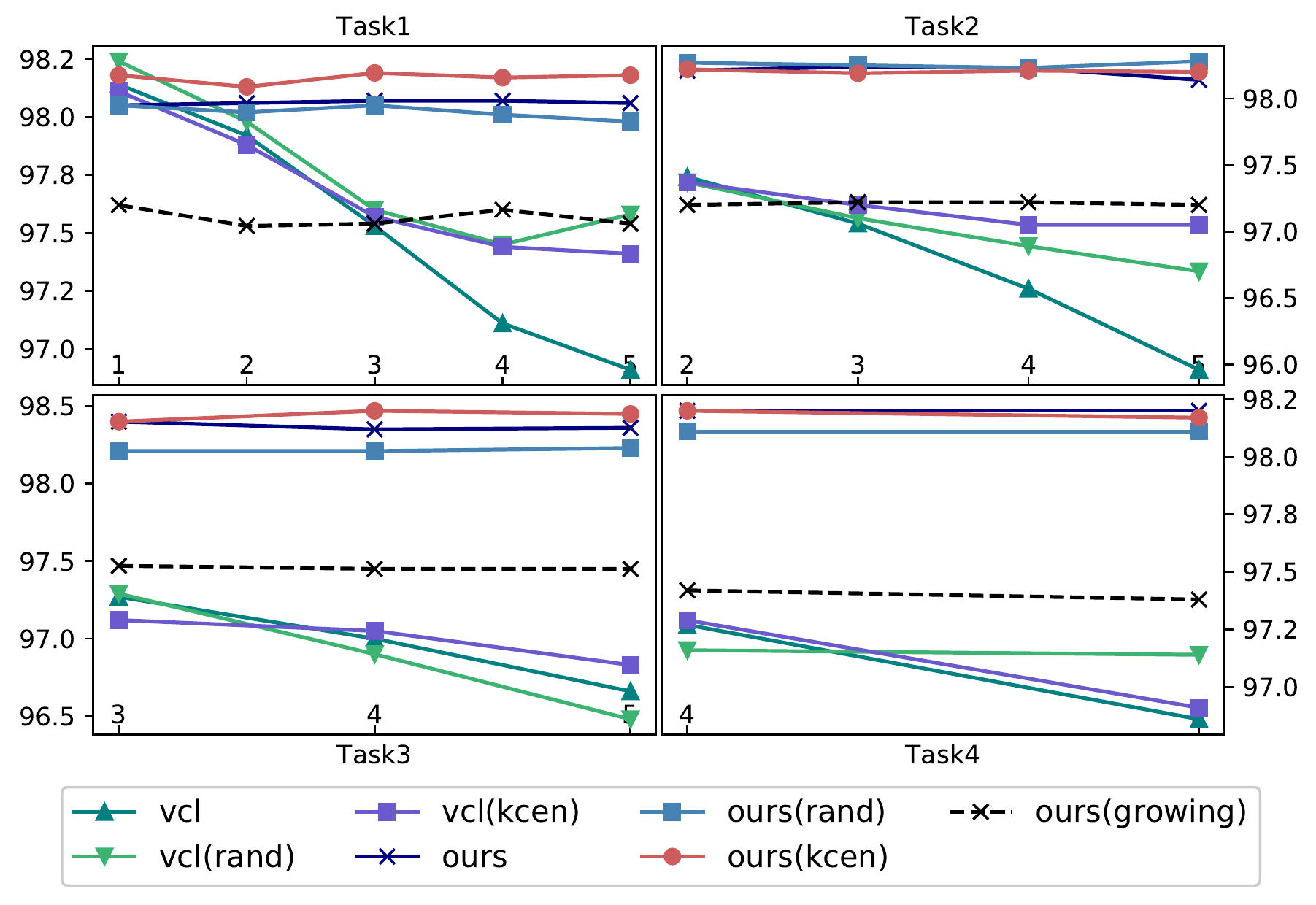}
    \vspace{-1em}
    \caption{ Comparison with VCL : Test accuracy of tasks 1, 2, 3, 4 (\textbf{left to right}) as newer tasks are observed for Permuted MNIST, random and k-center are methods used for coreset selection}
    \label{fig:fig3}
    \vspace{-1em}
    % \caption{}
\end{figure}

We use a feed-forward network (ReLU activations) with a single hidden layer having total budget of 200 nodes. For fair comparison, we use the same architecture for each of the baselines, except for DEN and RCL that grows the structure with two hidden layers. 
We also report results on some additional CL metrics \cite{cl_metrics}  in the Supplementary Material.
% \vspace{1mm}
\paragraph{Quantitative Results:} Fig.~\ref{fig:myfig}  shows the mean test accuracies on splitMNIST, notMNIST, permuted MNIST, and fashion MNIST as new tasks are observed. As Fig.~\ref{fig:myfig} shows, the average test accuracy of our method (without as well as with coresets) is better than the other baseline (here, we have used random point selection method for coresets). Moreover, the accuracy drops much more slowly than and other baselines showing the efficacy of our model in preventing catastrophic forgetting due to the adaptively learned structure.
% \ \\\ \\\ \\\ \\\ \\\ \\\ \\\ \\\ \\\ \\\ \\\ \\\ \\\ \\
%\vspace{-2em}
% \vspace{1.5mm}
In Fig.~\ref{fig:fig3}, we show the accuracy on individual tasks (for tasks 1-4) as new tasks arrive and compare specifically with VCL. In this case too, we observe that our method yields relatively stable individual task accuracies as compared to VCL. Also, as with VCL, using coreset was found to improve performance a bit. We also note that some of the old tasks' accuracies increases with training of new tasks which shows the presence of backward transfer, which is another desideratum of CL. We also report the performance with dynamic expansion of network initialized to 50 hidden units; it performs slightly worse than the truncation-based method but better than the other methods.

One further observation is that, for VCL, while the individual test accuracies improve with more training on each task, the overall performance across all tasks drops gradually, possibly since more training on a single task adapts the model more specifically for that task, leading to forgetting of the previous tasks. Our model, on the other hand, was found to be immune to over-training, since each task learns its own sparse subset of parameters.

\begin{figure*}[t!]
% \vspace{-1em}
    \centering
    \includegraphics[width=1.1\linewidth,trim={25em ,30em ,25em, 0em},clip]{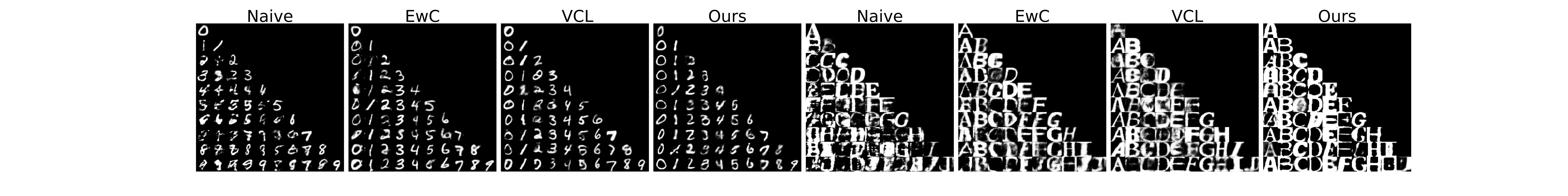}
\vspace{-3em}
    \caption{ Sequential generation for MNIST and notMNIST datasets (Supp. Material contains more illustrations and zoomed-in versions).}
    \label{fig:fig4}
    \vspace{-1em}
    % \caption{}
\end{figure*}

\subsection{Unsupervised Continual Learning}
We next evaluate our model for generative tasks under CL setting. For this evaluation, compare our model with existing approaches such as Na\"ive, EwC \cite{ewc} and VCL \cite{vcl}. We do not include other methods mentioned in supervised setup as their implementation does not incorporate generative modeling.

We perform continual learning experiments for deep generative models using a VAE style network. We consider two datasets, MNIST and notMNIST (small). For MNIST, the tasks are sequence of single digit generation from 0 to 9. Similarily, for notMNIST we define each task as one character generation from A to J.

Note that, unlike VCL and other baselines where all tasks have separate encoder and a shared decoder with separate head for latent dimension, as we discuss in Sec.~\ref{sec:unsup}, our model uses a shared encoder for all tasks, but with task-specific masks for each encoder (cf., Fig.~\ref{fig:schema} (b)). This enables transfer of knowledge while the task-specific mask effectively prevent catastrophic forgetting.
\begin{figure}[!htbp]\label{fig:ibp}
    \centering % <-- added
    % \begin{subfigure}{\linewidth}
    % \centering % <-- added
    \includegraphics[width=0.7\linewidth,trim={0em 0em 0em 0em},clip]{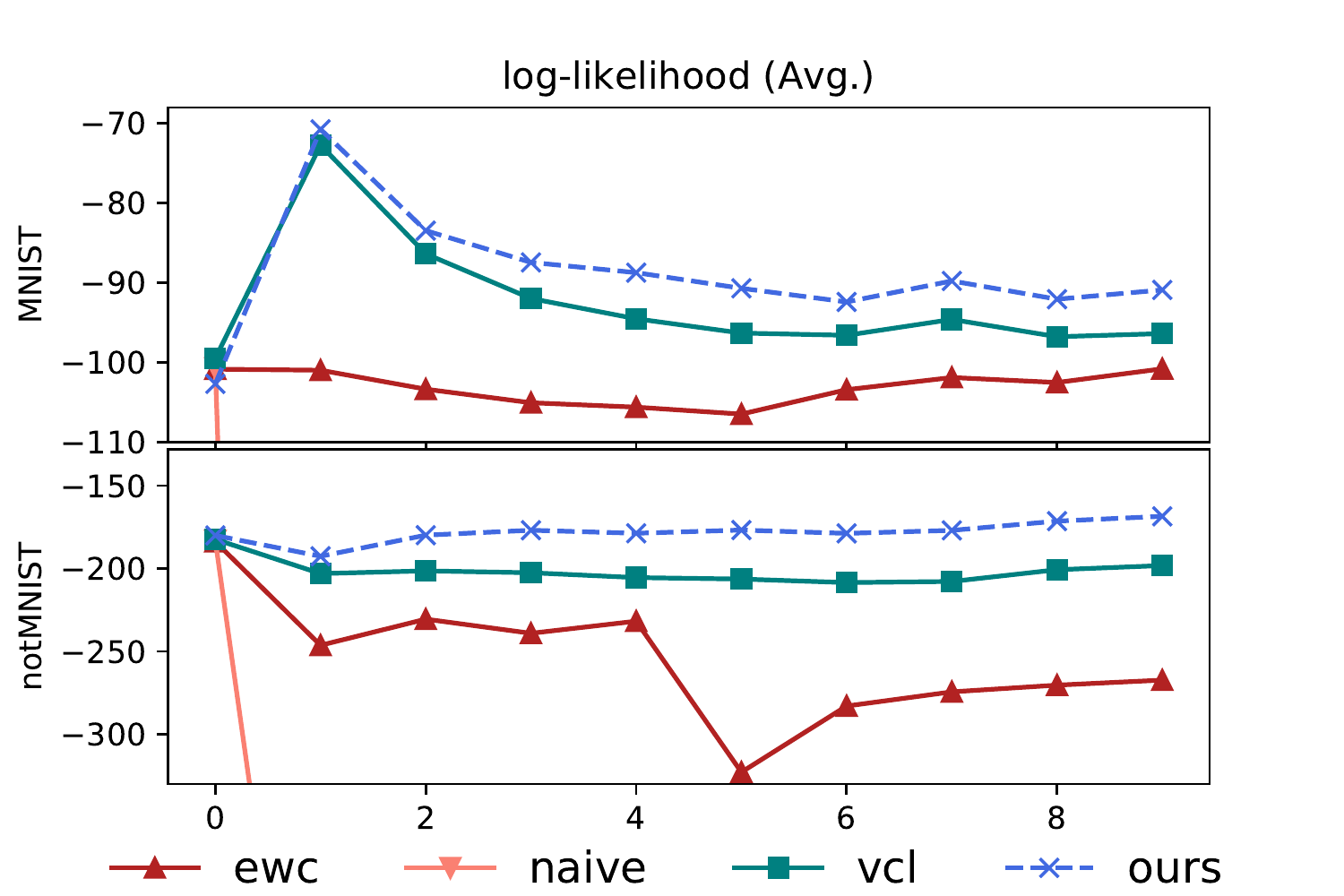}
    % \end{subfigure}
    \vspace{-3.0mm}
    \caption{\small{Log-likelihoods of benchmarks on generative tasks.}}
    \vspace{-3.0mm}
    \label{fig:logliks}
\end{figure}
\subsubsection{Performance evaluation}
%Suppose we have $T$ tasks arriving sequentially. To gauge the effectiveness of our model towards preventing catastrophic forgetting, we report the average log-likelihood of task $t$ after learning each of its subsequent tasks ($t, t+1, t+2, \dots T$).

\textbf{Generation:} As shown in Fig ~\ref{fig:logliks}, the modeling innovation we introduce for the unsupervised setting, indeed results in much improved log-likelihood on held-out sets. We also observe that the quality of generated samples in Fig ~\ref{fig:fig4} does not deteriorate as compared to other baselines as more and more tasks are encountered. In each individual figure in Fig ~\ref{fig:fig4}, each row represents the generated samples from all previously seen tasks and the current task. This shows that our model can efficiently perform generative modeling by reusing subset of networks and creating minimal number of nodes for each task.

\textbf{Representation Learning:} We also perform an experiment to assess the quality of the \emph{unsupervisedly} learned representation by our unsupervised continual learning approach. For this experiment, we use the learned representations to train a classification model. Due to space limitation, the details of this experiment are provided in the Supplementary Material.

\subsection{Some Structural Observations}
An appealing aspect of our work is that, the results reported above, which are competitive with the state-of-the-art, are achieved with a very sparse neural network structures learnt by the model, which we analyze qualitatively here (the Supplementary Material shows some examples of network structures learnt by our model).

\begin{figure}[h!]
    \centering
    \includegraphics[width=\linewidth,trim={3em 0em 3em 0em},clip]{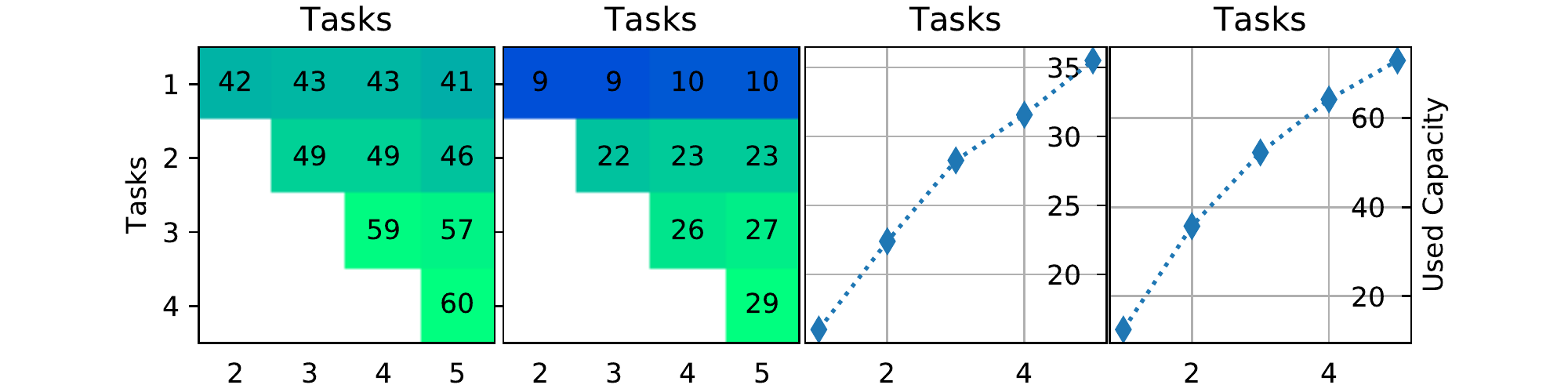}
    \begin{minipage}{0.45\linewidth}
        \centering
        (a)
    \end{minipage}
    \begin{minipage}{0.45\linewidth}
        \centering
        (b)
    \end{minipage}
    \vspace{-1em}
    \caption{\small{Percentage of weight sharing across tasks (\textbf{a}), percentage of network filled or used after learning k tasks (\textbf{b}), the first and third plots are from notMNIST results and the second and fourth plots are fromr permuted MNIST results}}
    \vspace{-1mm}
    \label{fig:figshare}
\end{figure}

Further, as expected by proposed model, as shown in Fig.~\ref{fig:figshare} (\textbf{b}), the IBP prior enforces the weights to be concentrated mainly on the first few nodes, and the structure results in a sparse network.  For both notMNIST and Permuted MNIST datasets, a maximum of around 15\% incoming connections are used at most for the first task. 

Another important observation (as shown in Fig.~\ref{fig:figshare} (\textbf{a})) is the percentage of weights that are being shared between different tasks and how the number of active weights vary across different tasks based on their similarities. Qualitatively, it appears that most newer tasks tend to allocate fewer weights and yet perform well, implying effective forward transfer. One can also easily observe that the weight sharing between similar tasks like those in notMNIST is a much higher than that of non-similar tasks such as permuted MNIST. This seemingly leads to the hypothesis that a single binary mask common for all tasks is sufficient. Experimental
observations, however, dispel such a belief by showing drastic degradation in performance.
We therefore conclude that although a new task tend to share weights learnt by old tasks, the new connections that it creates are indispensable for its performance. Intuitively, the more unrelated a task is to previously seen ones, the more new connections it will make, thus reducing \textit{negative transfer} (an unrelated task adversely affecting other tasks) between tasks.

%Figure ~5 demonstrates the learned patterns exhibited by the active weights for each of the first 5 hidden nodes. Note the similarity of the patterns across rows(tasks), which show the promise of our model for retaining the learned information. 

\vspace{-1em}

\section{Conclusion}
We have successfully unified structure learning in neural networks with their variational inference in the setting of continual learning, demonstrating competitive performance with state-of-the-art models on both discriminative (supervised) and generative (unsupervised) learning problems. It would also be interesting to extend this idea to more sophisticated network architectures such as convolutional or residual networks, possibly by also exploring improved approximate inference methods. Another interesting extension would be for semi-supervised continual learning. Adapting other sparse Bayesian structure learning methods, e.g. \cite{horseshoe} to the continual learning setting is also a promising avenue.  Adapting the \textit{depth} of the network is a more challenging endeavour that might also be undertaken. We leave these extensions for future work.

\renewcommand\thesubsection{\Alph{subsection}}
\twocolumn[
\icmltitle{Supplementary Material \\Bayesian Structure Adaptation for Continual Learning}
\vskip 0.3in
]
\appendix
% \section{Appendix}
\label{submission}
\subsection{Data}
The data sets used in our experiments with train test split information are listed in table given below. MNIST\footnote{MNIST : \urlx{http://yann.lecun.com/exdb/mnist/}} dataset comprises $28 \times 28$ monochromatic images consisting of handwritten digits from 0 to 9. notMNIST\footnote{
notMNIST : \urlx{http://commondatastorage.googleapis.com/books1000/notMNIST_small.tar.gz}} dataset comprises of glyph's of letters A to J in different fonts formats with similar configuration as MNIST. fashion MNIST\footnote{
fashion MNIST : \urlx{https://github.com/zalandoresearch/fashion-mnist/}
} is also monochromatic comprising of 10 classes (T-shirt, Trouser, Pullover, Dress, Coat, Sandal, Shirt, Sneaker, Bag, Ankle boot) with similar to MNIST.

% \begin{center}

\begin{tabular}{cccc}
\hline
     Dataset & Classes & Training size & Test size\\
     \hline\\
     MNIST & 10 & 60000 & 10000\\
     notMNIST & 10 & 14974 & 3750\\
     fashionMNIST & 10 & 50000 & 20000\\
     \hline
\end{tabular}
% \end{center}
\subsection{Model Configurations}
For permuted MNIST, split MNIST, split notMNIST and fashion MNIST experiments, we use fixed architecture of network for all the models with single hidden layer of 200 units except for DEN (which grows structure dynamically) which used two hidden layers initialized to $256, 128$ units. 

The VCL implementation was taken directly from their official repository at
\url{https://github.com/nvcuong/variational-continual-learning}. For DEN we used the official implementation at \url{https://github.com/jaehong-yoon93/DEN}. IMM implementation was taken from \url{https://github.com/btjhjeon/IMM\_tensorflow}, RCL implementation was taken from \url{https://https://github.com/xujinfan/Reinforced-Continual-Learning}, For EwC we used HAT's official implementation at \url{https://github.com/joansj/hat}. For rest of the models, we used our own implementations.

\subsubsection{Supervised Continual Learning: Hyperparameter settings}
For all datasets, our model uses single hidden layer neural network with $200$ hidden units. For RCL~\cite{rcl} and DEN~\cite{den}, two hidden layers were used with initial network size of $256,128$ units, respectively. We adopt Adam optimizer for our model keeping a learning rate of $0.01$ for the IBP posterior parameters and $0.001$ for others; this is to avoid vanishing gradient problem introduced by sigmoid function. For selective finetuning, we use a learning rate of $0.0001$ for all the parameters. The temperature hyperparameter of the Gumbel-softmax reparameterization for Bernoulli gets annealed from 10.0 to a minimum limit of 0.25. The value of $\alpha$ is initialized to 30.0 for the initial task and maximum of the obtained posterior shape parameters for each of subsequent tasks. Similar to VCL, we initialize our models with maximum-likelihood training for the first task. For all datasets, we train our model for 5 epochs. We selectively finetune our model after that for 5 epochs. For experiments including coresets, we use a coreset size of 50. Coreset selection is done using random and $k$-center methods \cite{vcl}. For our model with dynamic expansion, we initialize our network with 50 hidden units.
 
 \subsubsection{Unsupervised Continual Learning: Hyperparameter settings}
 For all datasets, our model uses 2 hidden layers with $500,500$ units for encoder and symmetrically opposite for the decoder with a latent dimension of size $100$ units. For other approaches like Naive, EwC and VCL \cite{ewc,vcl}, we use task-specific encoders with 3 hidden layers of $500,500,500$ units respectively with latent size of $100$ units, and a symmetrically reversed decoder with last two layers of decoder being shared among all the tasks and the first layer being specific to each task. we use Adam optimizer for our model keeping the learning rate configuration similar to that of supervised setting. Temperature for gumbel-softmax reparametrization gets annealed from 10 to 0.25. We initialize encoder hidden layers $\alpha$ values as $40,40$, respectively, and symmetrically opposite in decoder for the first task. We update $\alpha$'s in similar fashion to supervised setting for subsequent tasks. For latent layers, we intialize $\alpha$ to $20$. For the unsupervised learning experiments, we did not use coresets.
 
 \begin{algorithm}[!tb]
\caption{Nonparametric Bayesian CL}\label{algo}
\begin{algorithmic}
\STATE {\bfseries Input:}{Initial Prior $p_0(\Theta)$}
\STATE Initialize the network parameters and coresets
\STATE Initialize : $p_{new} \leftarrow p_0(\Theta)$
    \FOR{$i=1$ {\bfseries to} $T$}
    \STATE Observe current task data $D_t$\;
    \STATE Update coresets (Appendix \ref{coresetApp})\;
    \STATE \textbf{Masked Training}\;
    
    \begin{ALC@g}
    \STATE $\mathcal L_t \leftarrow$ ELBO (Eq. 6) with prior $p_{new}$\;
    \STATE $\Theta_t \leftarrow \arg \min \mathcal L_t$\;
    \end{ALC@g}
    \STATE \textbf{Selective Finetuning}\;
    \begin{ALC@g}
    \STATE Fix the IBP parameters and learned mask\;
    \STATE $\Theta_t \leftarrow \arg \min \mathcal L_t$\;
    \end{ALC@g}
    \STATE $p_{new} \leftarrow q_t(\Theta)$\;
    \STATE $p_{new} \leftarrow$ Mask($p_{new}$) using Eq 11\;
    \STATE Perform prediction for given test set..
    \ENDFOR
\end{algorithmic}
\end{algorithm}
\subsection{Coreset Method Explanation}\label{coresetApp}
As done in VCL~\cite{vcl}, for each task, a new coreset is produced by selecting a few points from the new task and the old coreset. Coreset selection can be done either through random selection or $K$-center greedy algorithm ~\cite{kcen}. Next, the posterior is decomposed as follows:
\[p(\theta|D_{1:t}) \propto p(\theta|D_{1:t} \char`\\ C_t)p(C_t|\theta) \approx \tilde q_t(\theta)p(C_t|\theta)\]
where, $q(\theta)$ is the variational posterior obtained using the current task training data, excluding the current coreset data. Applying this trick in a recursive fashion, we can write:
\begin{align*}
    p(\theta|D_{1:t} \char`\\ C_t) &= p(\theta|D_{1:t-1} \char`\\ C_{t-1})p(D_t \cup C_{t-1}\char`\\ C_t|\theta)\\
    &\approx \tilde q_{t-1}(\theta)p(D_t \cup C_{t-1}\char`\\ C_t|\theta)
\end{align*}
We then approximate this posterior using variational approximation as $\tilde q_t(\theta) = proj(\tilde q_{t-1}(\theta)p(D_t \cup C_{t-1}\char`\\ C_t|\theta))$
Finally a projection step is performed using coreset data before prediction as follows: $q_t(\theta) = proj(\tilde q_t(\theta)p(C_t|\theta))$. This way of incorporating coresets into coreset data before prediction tries to mitigate any residual forgetting. Algorithm \ref{algo} summarizes the training procedure for our model.

\subsection{Additional Inference Details}
Inference over parameters $\phi$ that involves a random or stochastic node $Z$ (i.e $Z\sim q_{\phi}(Z)$) cannot be done in a straightforward way, if the objective involves Monte Carlo expectation with respect that random variable $(\mathcal L = \mathbb E_{q_{\phi}{z}}(L(z))))$. This is due to the inability to back-propagate through a random node. To overcome this issue, \cite{kingma2013auto} introduced the reparametrization trick. This involves deterministically mapping the random variable $Z = f(\phi, \epsilon)$ to rewrite the expectation in terms of new random variable $\epsilon$, where $\epsilon$ is now randomly sampled instead of $Z$ (i.e $\mathcal L = \mathbb E_{q{\epsilon}}[L(\epsilon, \phi)]$). In this section, we discuss some of the reparameterization tricks we used.
\subsubsection{Gaussian distribution Reparameterization}
The weights of our Bayesian nueral network are assumed to be distributed according to a Gaussian with diagonal variances (i.e $V_{k}\sim \mathcal N(V_{k}|\mu_{V_{k}}, \sigma^{2}_{V_{k}})$). We reparameterize our parameters using location-scale trick as:
\[V_{k} = \mu_{V_{k}} + \sigma_{V_{k}}\times \epsilon, \quad \epsilon \sim \mathcal N(0, I)\]
where $k$ is the index of parameter that we are sampling. Now, with this reparameterization, the gradients over $\mu_{V_{k}}, \sigma_{V_{k}}$ can be calculated using back-propagation.
\subsubsection{Beta distribution Reparameterization}
The beta distribution for parameters $\nu$ in the IBP posterior can be reparameterized using Kumaraswamy distribution \cite{kumara}, since Kumaraswamy distribution and beta distribution are identical if any one of rate or shape parameters are set to 1. The Kumaraswamy distribution is defined as
$p(\nu;\alpha,\beta) = \alpha\beta \nu^{\alpha-1}(1-\nu^{\alpha})^{\beta - 1}$ which can be reparameterized as:
\[\nu = (1-u^{1/\beta})^{1/\alpha}, \quad u\sim U(0,1)\]
where $U$ represents a uniform distribution. The KL-Divergence between Kumaraswamy and  beta distributions can be written as:
\begin{align}
    KL(q(\nu; a,b)||&p(\nu; \alpha, \beta)) = \frac{a-\alpha}{a}\left(-\gamma -\Psi(b) -\frac{1}{b} \right)\notag \\
    &+ \log{ab} + \log(B(\alpha, \beta)) -\frac{b}{1-b}\notag\\
    &+ (\beta-1)b\sum_{m=1}^{\infty}\frac{1}{m+ab}B(\frac{m}{a},b)\label{eq:kumakl}
\end{align}
where $\gamma$ is the Euler constant, $\Psi$ is the digamma function and B is the beta function. As described in \cite{kumara}, we can approximate the infinite sum in Eq.\ref{eq:kumakl} with a finite sum using first 11 terms.
\subsubsection{Bernoulli distribution Reparameterization}
For Bernoulli distribution over mask in the IBP posterior, we employ the continuous relaxation of discrete distribution as proposed in Categorical reparameterization with Gumbel-softmax \cite{gumbsoft}, also known as the CONCRETE \cite{concrete} distribution. We sample a concrete random variable from the probability simplex as follows:
\[B_{k} = \frac{\exp((\log(\alpha_k) + g_{k})/\lambda)}{\sum^K_{i=1}\exp((\log(\alpha_i) + g_{i})/\lambda)},\quad g_k\sim G(0,1)\]
where, $\lambda \in (0,\infty)$ is a temperature hyper-parameter, $\alpha_k$ is posterior parameter representing the discrete class probability for $k^{th}$ class and $g_{k}$ is a random sample from Gumbel distribution $G$. For binary concrete variables, the sampling reduces to the following form:
\[Y_{k} = \frac{\log{(\alpha_k)} + \log{(u_k/(1-u_k))}}{\lambda},\quad u\sim U(0,1)\]
then, $B_k = \sigma(Y_k)$
where $\sigma$ is sigmoid function and $u_k$ is sample from uniform distribution U. To guarantee a lower bound on the ELBO, both prior and posterior Bernoulli distribution needs to be replaced by concrete distributions. Then the KL-Divergence can be calculated as difference of log density of both distributions. The log density of concrete distribution is given by:
\begin{align*}
    \log q(B_k; \alpha, \lambda) &= \log{(\lambda)} - \lambda Y_{k} + \log{\alpha_k}\\
&-2\log{(1+\exp{(-\lambda Y_k + \log{\alpha_k})})}
\end{align*}
With all reparameterization techniques discussed above, we use Monte Carlo sampling for approximating the ELBO with sample size of 10 while training and a sample size of 100 while at test time.

\subsection{IBP Hyperparameter $\alpha$}
In this section, we discuss the approach to tune the IBP prior hyperparameter $\alpha$. As discussed earlier, we found that using a sufficiently large value of $\alpha$ without tuning performs reasonably well in practice. However, we experimented with other alternatives as well. For example, we tried adapting $\alpha$ with respect to previous posterior as $\alpha = max(\alpha, max(a_\nu))$ for each layer, where $a_\nu$ is Beta posterior shape parameter. Several other considerations can also be made regarding its choice.

\subsubsection{Scheduling across tasks} Intuitively, $\alpha$ should be incremented for every new task according to some schedule. Information about task relatedness can be helpful in formulating the schedule. Smaller increments of $\alpha$ discourages creation of new nodes and encourages more sharing of already existing connections across tasks. 

\subsubsection{Learning $\alpha$} Although not investigated in this work, one viable alternative to choosing $\alpha$ by cross-validation could be to learn it. This can be accommodated into our variational framework by imposing a gamma prior on $\alpha$ and using a suitably parameterized gamma variational posterior. The only difference in the objective would be in the KL terms: the KL divergence of $\bm v$ will then also have to estimated by Monte Carlo approximation (because of dependency on $\alpha$ in the prior). Also, since gamma distribution does not have an analytic closed form KL divergence, the Weibull distribution can be a suitable alternative~\cite{zhang2018whai}.

\subsection{Additional Results: Supervised Continual Learning}
In this section, we provide some additional experimental results for supervised continual learning setup. Table \ref{tab:mytab} shows final mean accuracies over 5 tasks with deviations, obtained by all the approaches on various datasets. It also shows that our model performs comparably or better than the baselines.
\subsubsection{Learned Network Structures}
In this section, we analyse the network structures that were learned after training our model.
\begin{figure}[!htpb]
    \centering
    \includegraphics[width=0.9\linewidth,trim={1em 6em 1em 5.5em},clip]{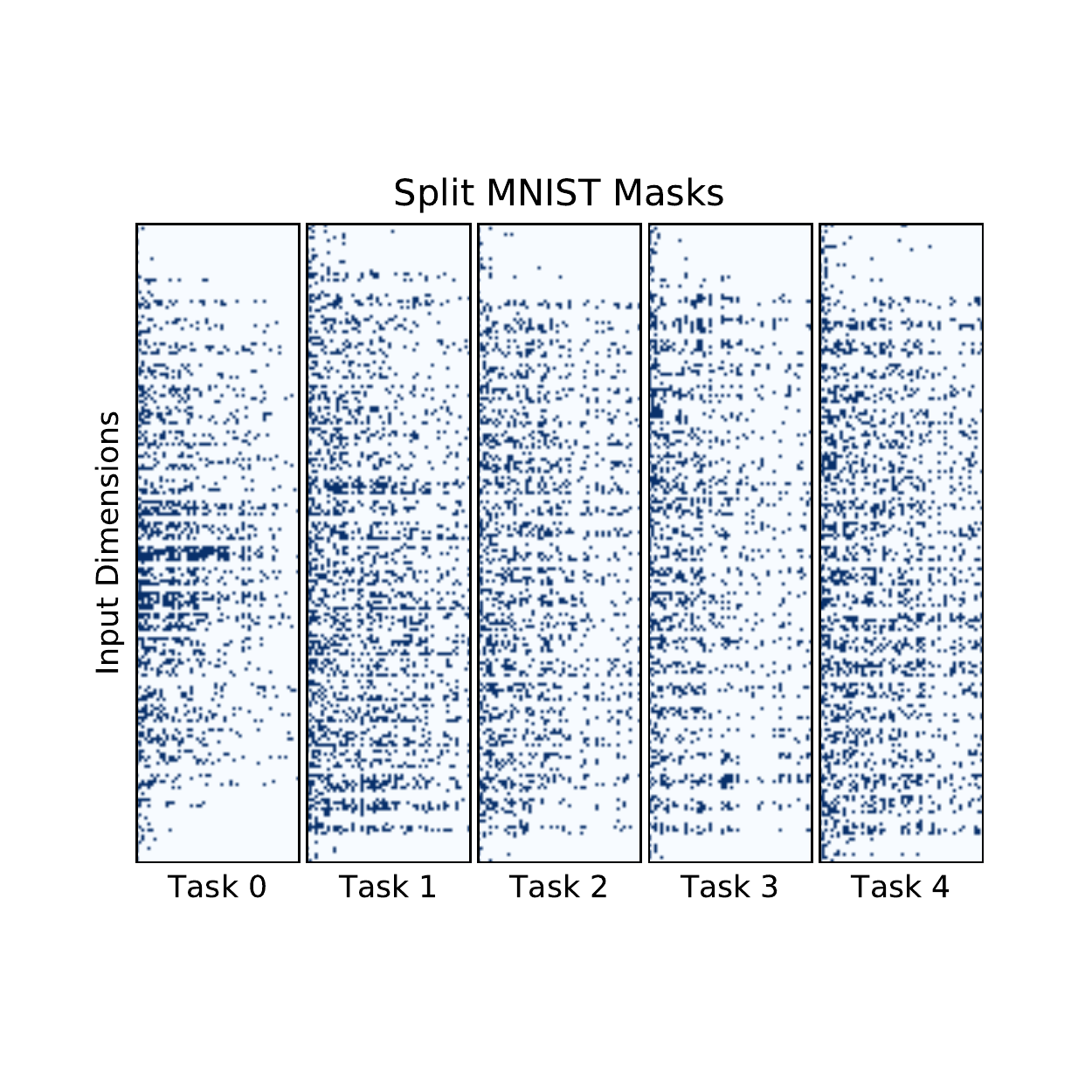}
    \caption{Learned masks for input to first hidden layer weights on splitMNIST dataset. Darker color represent active weights.}
    \label{fig:splitmasks}
\end{figure}
As we can see in Fig. \ref{fig:splitmasks}, the masks are captured on the pixel values where the digits in MNIST datasets have high value and zeros elsewhere which represents that our models adapts with respect to data complexity and only uses those weights that are required for the task. Due to the use of the IBP prior, the number of active weights tends to shrink towards the first few nodes of the first hidden layer. This observation enforces that our idea of using IBP prior to learn the model structure based on data complexity is indeed working. Similar behaviour can be seen in notMNIST and fashionMNIST in Fig.  \ref{fig:othermasks}. On the other hand Fig~\ref{fig:filled2} (left) shows the sharing of weights between subsequent tasks of different datasets. It can be observed that the tasks that are similar at input level of representation have more overlapping/sharing of parameters (e.g split MNIST) in comparison to those that are not very similar (e.g permuted MNIST). It also shows Fig~\ref{fig:filled2} (right) that the amount of total network capacity used by our model differs for each task, which shows that complex tasks require more parameters as compared to easy tasks. 
\begin{table*}[!tb]
% \vspace{-1em}
\centering
  \begin{tabular}{p{3.7cm} p{2.8cm} p{2.8cm} p{2.8cm} p{2.8cm}}
 \toprule
 Method & split MNIST & notMNIST & permuted MNIST & fashion MNIST\\
 \midrule
 Na\"ive & 79.615$\pm$0.7 & 72.339$\pm$0.8 & 90.090$\pm$0.4 & 79.319$\pm$0.6\\ 
 Rehearsal & 99.102$\pm$0.3 & 95.203$\pm$0.5 & 97.565$\pm$0.3& 97.981$\pm$0.3\\
 EwC & 81.530$\pm$0.4 & 90.297$\pm$0.6 & 95.392$\pm$0.5& 86.577$\pm$0.4\\
 IMM (mode) & 92.206$\pm$0.6 & 84.442$\pm$0.4 & 96.433$\pm$0.5 & 88.765$\pm$0.4\\
 VCL & 98.952$\pm$0.3 & 93.732$\pm$0.3 & 97.353$\pm$0.3& 97.970$\pm$0.2\\
 VCL(coreset) & 98.731$\pm$0.4 & 94.993$\pm$0.2 & 97.464$\pm$0.3& 98.154$\pm$0.3\\
 DEN & 99.779$\pm$0.1 & 96.485$\pm$0.3 & 97.945$\pm$0.2& 98.580$\pm$0.3\\
 RCL & 99.768$\pm$0.1 & 96.722$\pm$0.2 & 98.005$\pm$0.2& 98.698$\pm$0.2\\
 Ours & 99.819$\pm$0.1 & \textbf{97.152}$\pm$0.2 & \textbf{98.180}$\pm$0.2& 98.986$\pm$0.2\\
 Ours(coreset) & \textbf{99.834}$\pm$0.1 & 97.061$\pm$0.2 & 98.163$\pm$0.3& \textbf{98.990}$\pm$0.2\\
 \bottomrule
 \end{tabular}
 \caption{Comparison of final mean accuracies on test set obtained using different methods. Deviations are rounded to 1 decimal place}\label{mytable}
\label{tab:mytab}
\end{table*}
Since the network size is fixed, the amount of network usage for all previous tasks tends to converge towards 100 percent. This promotes parameter sharing but also introduces forgetting, since the network is forced to share parameters and is not able to learn new nodes. 

\begin{figure}[!htpb]
    \centering
    \includegraphics[width=0.85\linewidth,trim={1em 6em 1em 5.5em},clip]{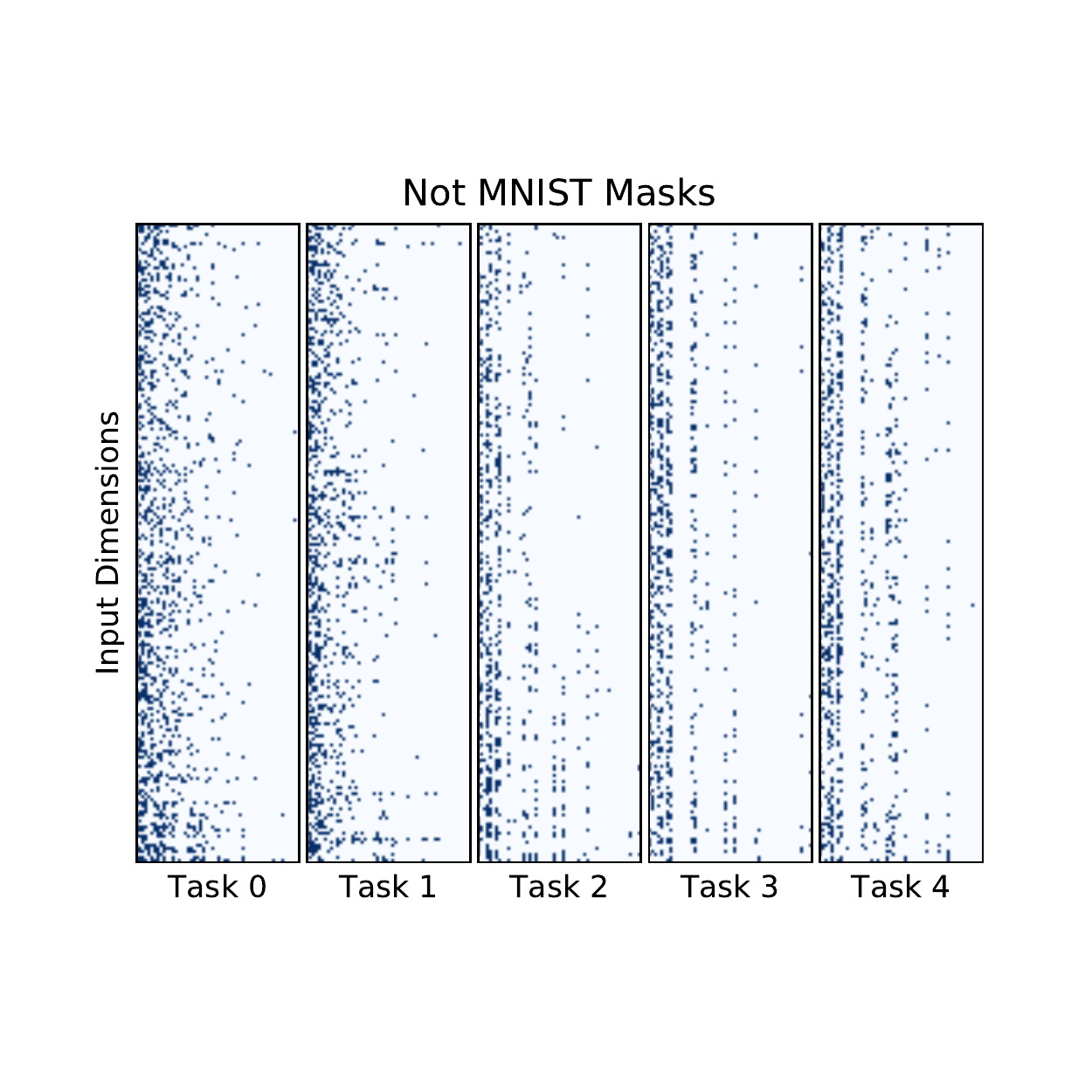}
    \includegraphics[width=0.85\linewidth,trim={1em 6em 1em 5.5em},clip]{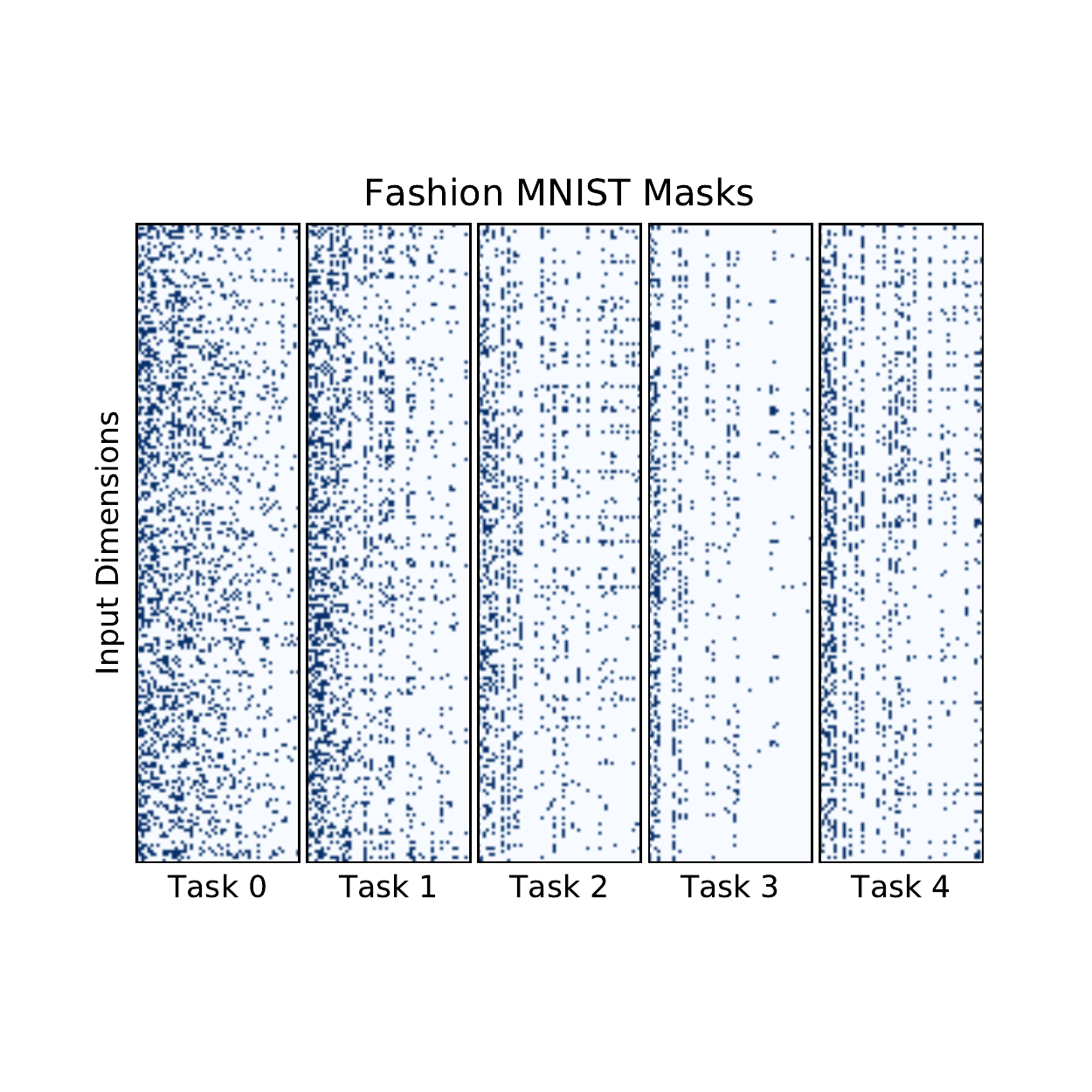}
    \caption{Learned masks for input to first hidden layer weights on  notMNIST(\textbf{top}) and fashion MNIST(\textbf{bottom}) datasets. Darker color represent active weights.}
    \label{fig:othermasks}
\end{figure}
\subsubsection{Other Metrics}
We quantified and observed the forward and backward transfer of our and VCL model, using the three metrics given in \cite{cl_metrics} on Permuted MNIST dataset as follows:
\subparagraph{Accuracy}is defined as the overall model performance averaged over all the task pairs as follows:
\[Acc = \frac{\sum_{i\geq j}R_{i,j}}{\frac{N(N-1)}{2}}\]
where, $R_{i,j}$ is obtained test classification accuracy of the model on task $t_j$ after observing the last sample from task $t_i$.
\begin{figure}[!htpb]
    \centering
    \includegraphics[width=0.7\linewidth,trim={0em 6em 1em 6em},clip]{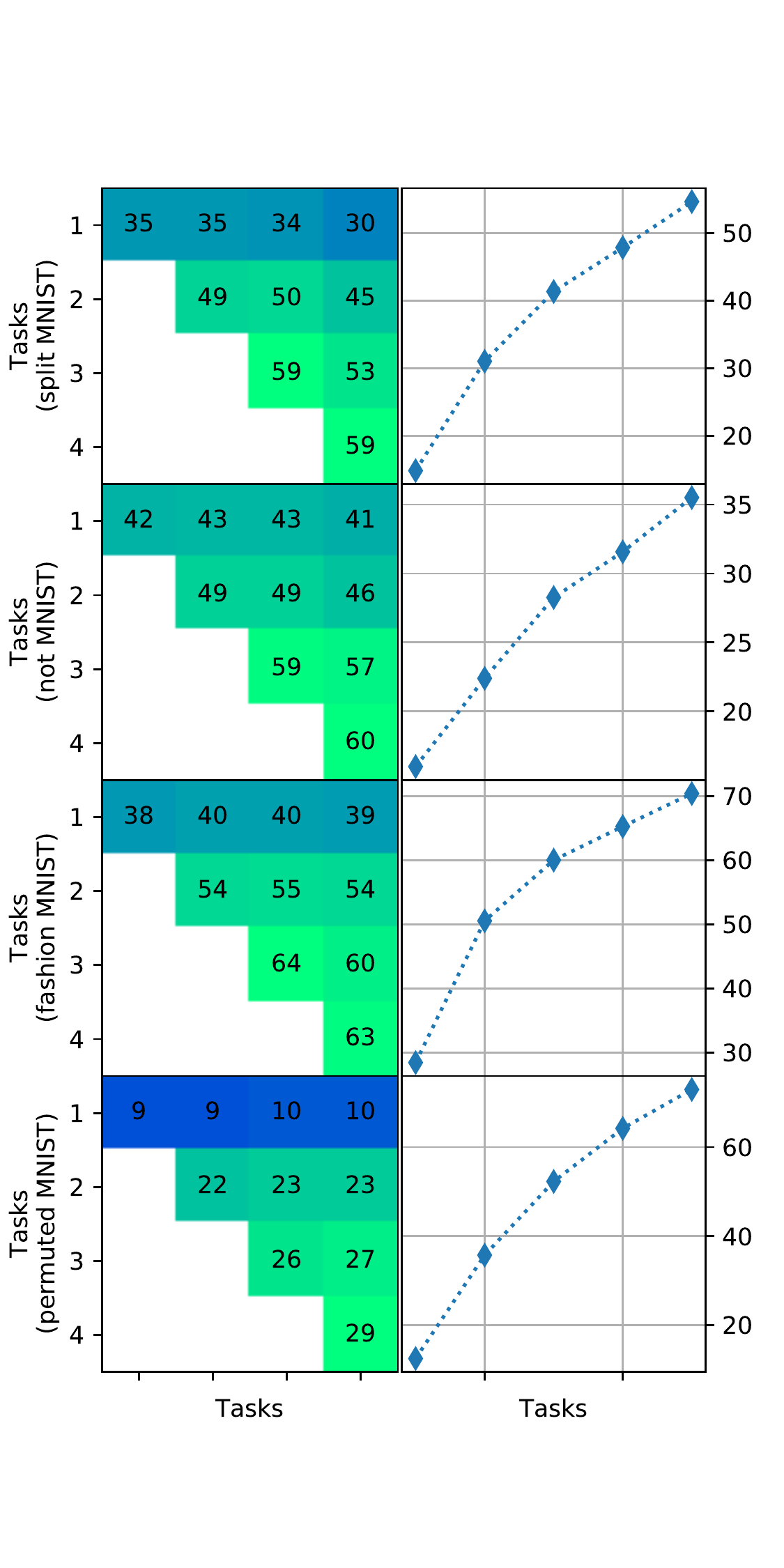}
    \caption{Percentage weight sharing between tasks (\textbf{left}), percentage of network capacity already used by previous tasks(\textbf{right}).}
    \label{fig:filled2}
\end{figure}
\subparagraph{Forward Transfer}
is the ability of previously learnt task to perform on new task better and is give by:
\[FWT = \frac{\sum_{i < j}^N R_{i,j}}{\frac{N(N-1)}{2}}\]
\subparagraph{Backward Transfer}
is the ability of newly learned task to affect the performance of previous tasks. It can be defined as:
\[BWT = \frac{\sum_{i=2}^N \sum_{j=1}^{i-1} (R_{i,j} - R_{j,j})}{\frac{N(N-1)}{2}}\]
We compare our model with VCL and other baselines over these three metrics in Table~\ref{tab:mytab2}.
\begin{table}[!htpb]
% \vspace{-1em}
   \begin{tabular}{p{2.0cm} p{1.3cm} p{1.3cm} p{1.55cm}}
 \toprule
 Method & Acc & FWT & BWT \\
 \midrule
 Naive & $90.090$ & $0.1$ & $-3.60e^{-2}$\\
 EwC & $95.392$ & $0.1$ & $-1.90e^{-2}$\\
 Rehearsal & $97.565$ & $0.1$ & $+1.30e^{-4}$\\
 VCL & $97.353$ & $0.1$ & $-4.00e^{-3}$\\
 Ours & $98.180$ & $0.1$ & $+1.33e^{-5}$\\
 \bottomrule
 \end{tabular}
 \caption{Comparison on other metrics for permuted MNIST dataset}
\label{tab:mytab2}
\end{table}
We can observe that backward transfer for our model is more as compared to most baselines, which shows that our approach has suffers from less forgetting as well. On the other hand forward transfer seems to give close to random accuracy (0.1) which is due to the fact that the model is not trained on the correct class labels and is asked to predict the correct label. So this metric is not very useful here; an alternative would be to train a linear classifier on the representations that are learned after each subsequent tasks for future task. 
\subsection{Unsupervised Continual Learning}
Here we describe the complete generative model for our unsupervised continual learning approach. The generative story for unsupervised setting can be written as follows (for brevity we have omitted the task id $t$):
\begin{align*}
    \bm B^{l} &\sim IBP(\alpha)\\
    \bm V_{d,k}^{l} &\sim \mathcal N(0,\sigma_0^2)\\
    \bm W^{l} &= \bm B^{l}\odot \bm V^{l}\\
    \bm W^{out}_{d,k} &\sim \mathcal N(0,\sigma_0^2)\\
    \bm Z_n &\sim \mathcal N(\mu_z,\sigma_z^2)\\
    \bm X_n \sim Bernoulli&( \sigma(\bm W^{out}\bm \phi_{NN}(\bm W, \bm Z_n)))
\end{align*}
where, $\mu_z, \sigma_z^2$  are prior parameters of latent representation; they can either be fixed or learned, and $\sigma$ is the sigmoid function. The stick-breaking process for the IBP prior remains the same here as well. For doing inference here, once again we resort to structured mean-field assumption:
\[q(\bm Z, \bm V, \bm B, \bm v) = q(\bm Z|\bm B,\bm V,\bm \nu,\bm X) q(\bm V) q(\bm B | v) q(\bm v)\]
where, $q(\bm Z|\bm B,\bm V,\bm \nu,\bm X) = \prod_{n=1}^N\mathcal N(\mu_{\phi_{NN}}, \sigma^2_{\phi_{NN}})$, and $\phi_{NN}$ is IBP masked neural network used for amortization of Gaussian posterior parameters. Rest of variational posteriors are factorized in a similar way as in the supervised approach. Evidence lower bound calculation can done as explained in section 3.3.

\subsubsection{Additional Experimental Results for Unsupervised Continual Learning}
In this section, we show further results for unsupervised continual learning. 
Fig \ref{fig:genliks} shows, for MNIST and notMNIST datasets, how the likelihoods vary for individual tasks as subsequent tasks arrive. 
It can be observed that the individual task accuracies \begin{figure}[!htpb]
    \centering
    \includegraphics[width = 0.75\linewidth,trim={4em 7em 4em 7em},clip]{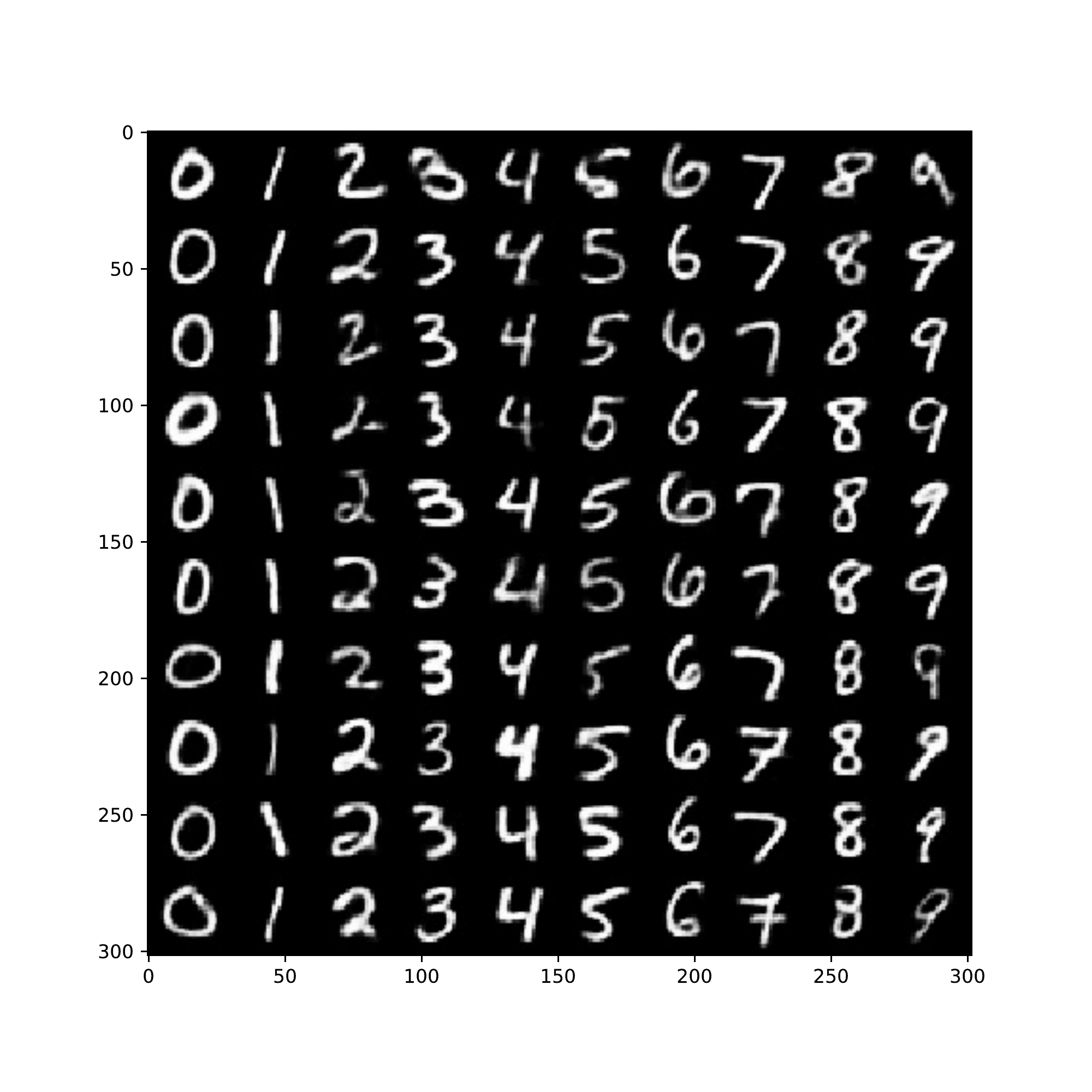}
    \includegraphics[width = 0.75\linewidth,trim={4em 7em 4em 7em},clip]{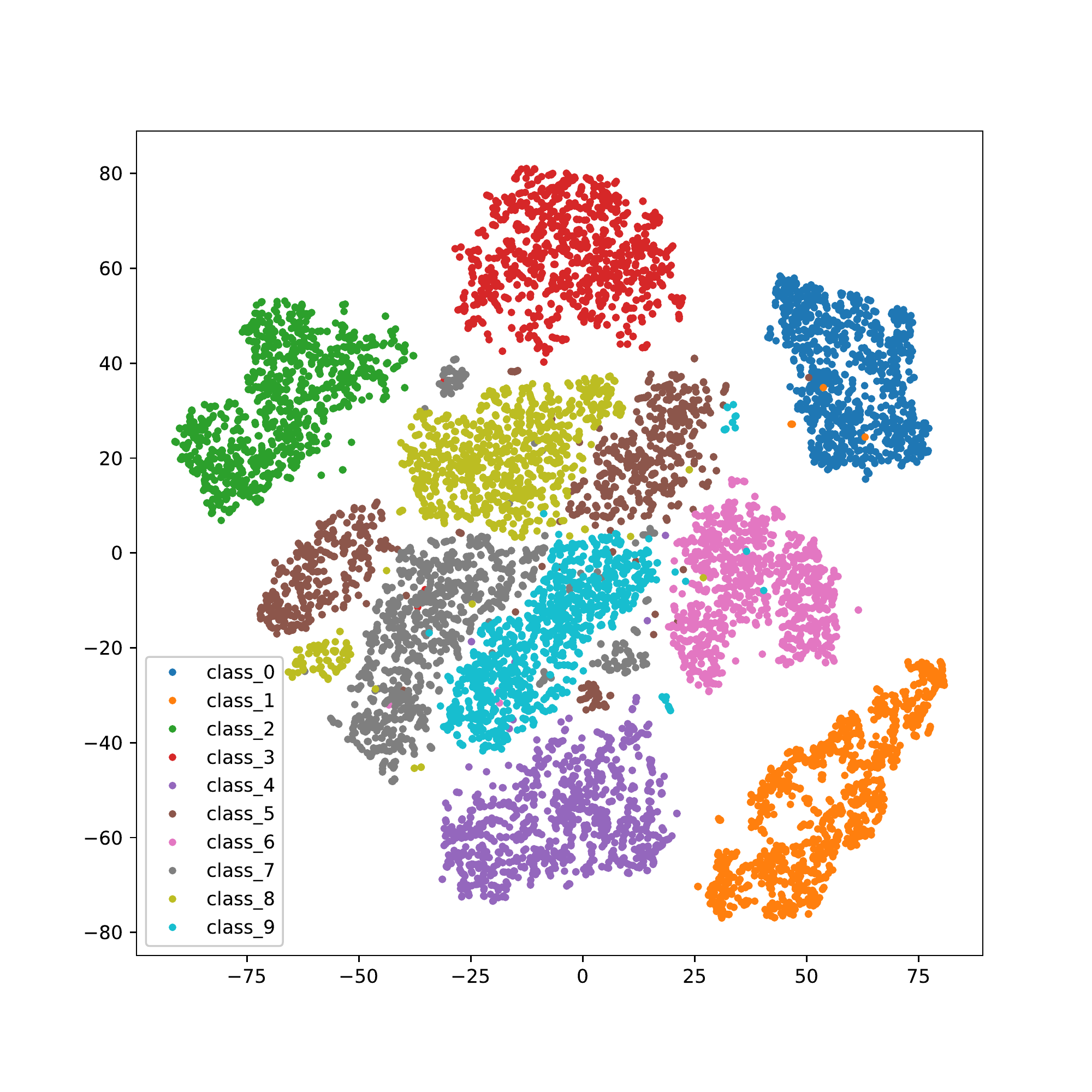}
    \caption{On MNIST dataset (\textbf{Top}) Reconstruction of images after all tasks have been observed. (\textbf{Bottom}) t-SNE plot of each class after all tasks have been observed.}
    \label{fig:recon_mnist}
\end{figure}
\begin{figure*}[!htpb]
    \centering
    \includegraphics[width=0.9\linewidth,trim={2em 2.5em 2em 2em},clip]{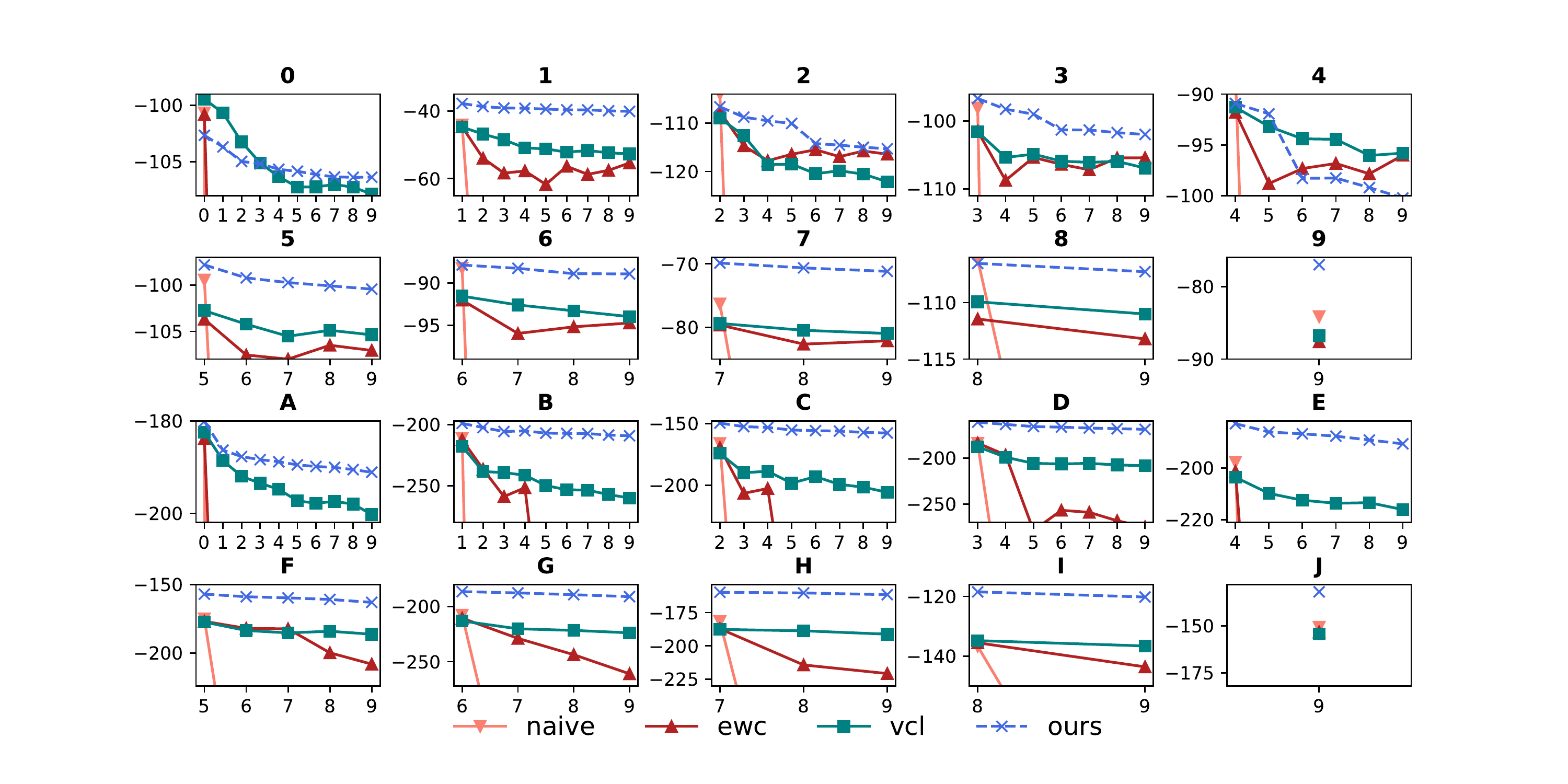}
    \caption{Generative Model : Test likelihood decays of individual tasks after subsequent tasks have been observed. (Top two) represents MNIST and (Bottom two) represents notMnist datasets.}
    \label{fig:genliks}
\end{figure*}
learned by our model are better than other baselines; this suggests that use of new weights when needed helps in retaining a better optima per task, and also the deterioration of our model is much less as compared to other model, representing effective protection against catastrophic forgetting.
Fig \ref{fig:recon_mnist} shows the reconstructed images of MNIST and also the t-SNE plot of latent codes our model produces. 
it can be observed that reconstruction quality is good despite heavy constraints on the model. 
\begin{figure}[!htpb]
    \centering
    \includegraphics[width = 0.8\linewidth,trim={4em 7em 4em 7em},clip]{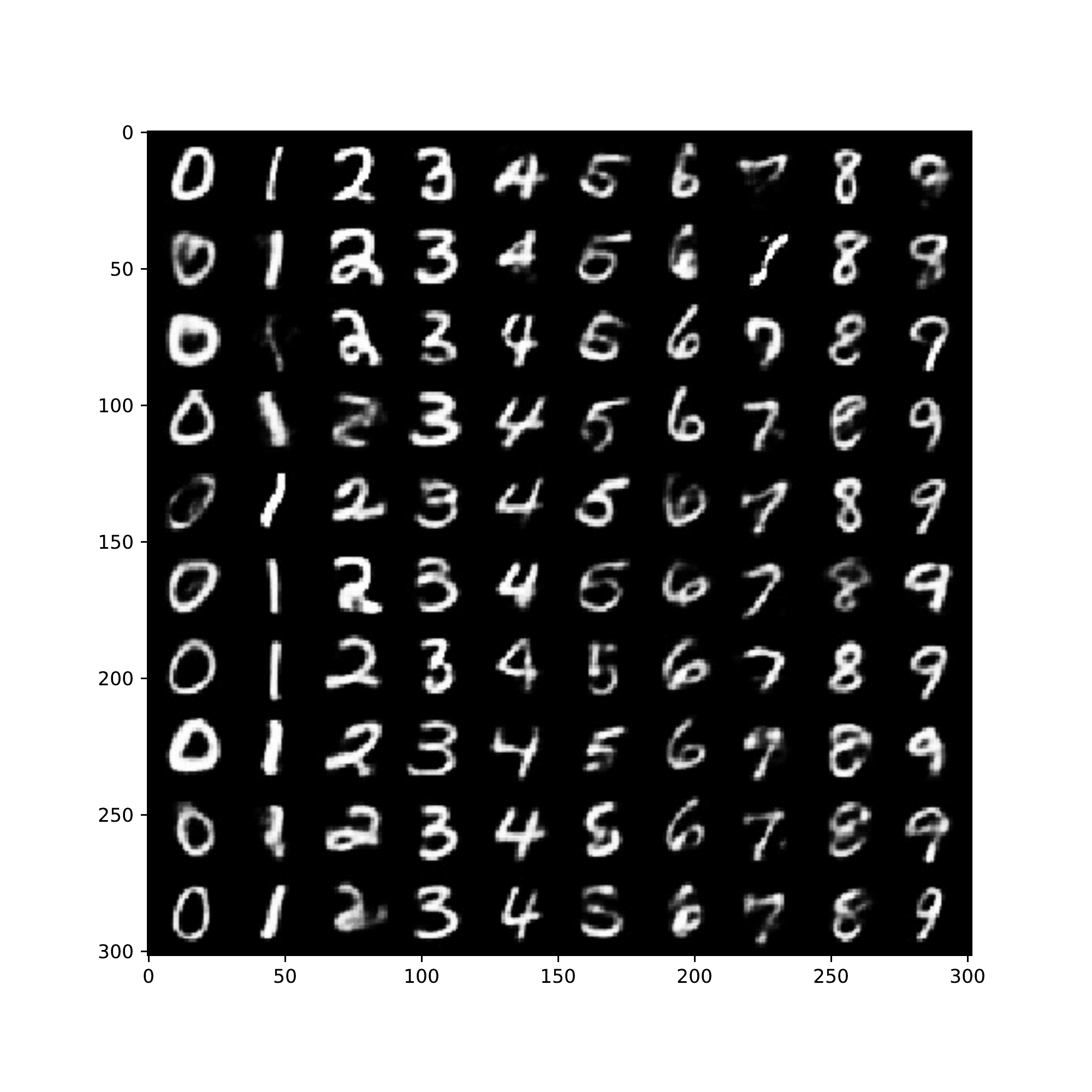}
    \caption{Generated samples on MNIST dataset after all tasks have been observed}
    \label{fig:gens_mnist}
\end{figure}
Fig \ref{fig:gens_mnist} shows the generated samples from the learned prior over latent space after all tasks are observed.
\begin{figure}[!htpb]
    \centering
    \includegraphics[width = 0.8\linewidth,trim={4em 7em 4em 7em},clip]{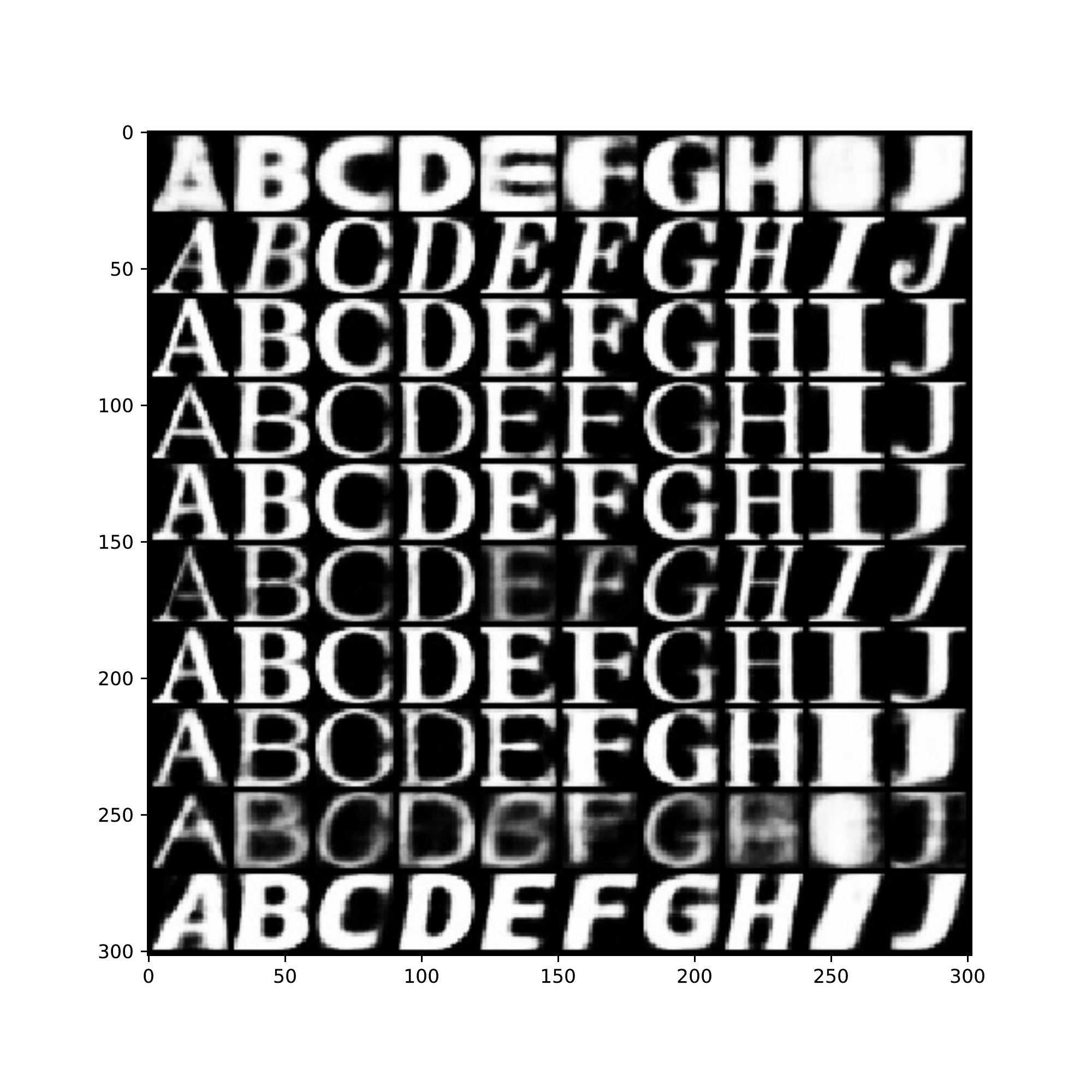}
    \includegraphics[width = 0.8\linewidth,trim={4em 7em 4em 7em},clip]{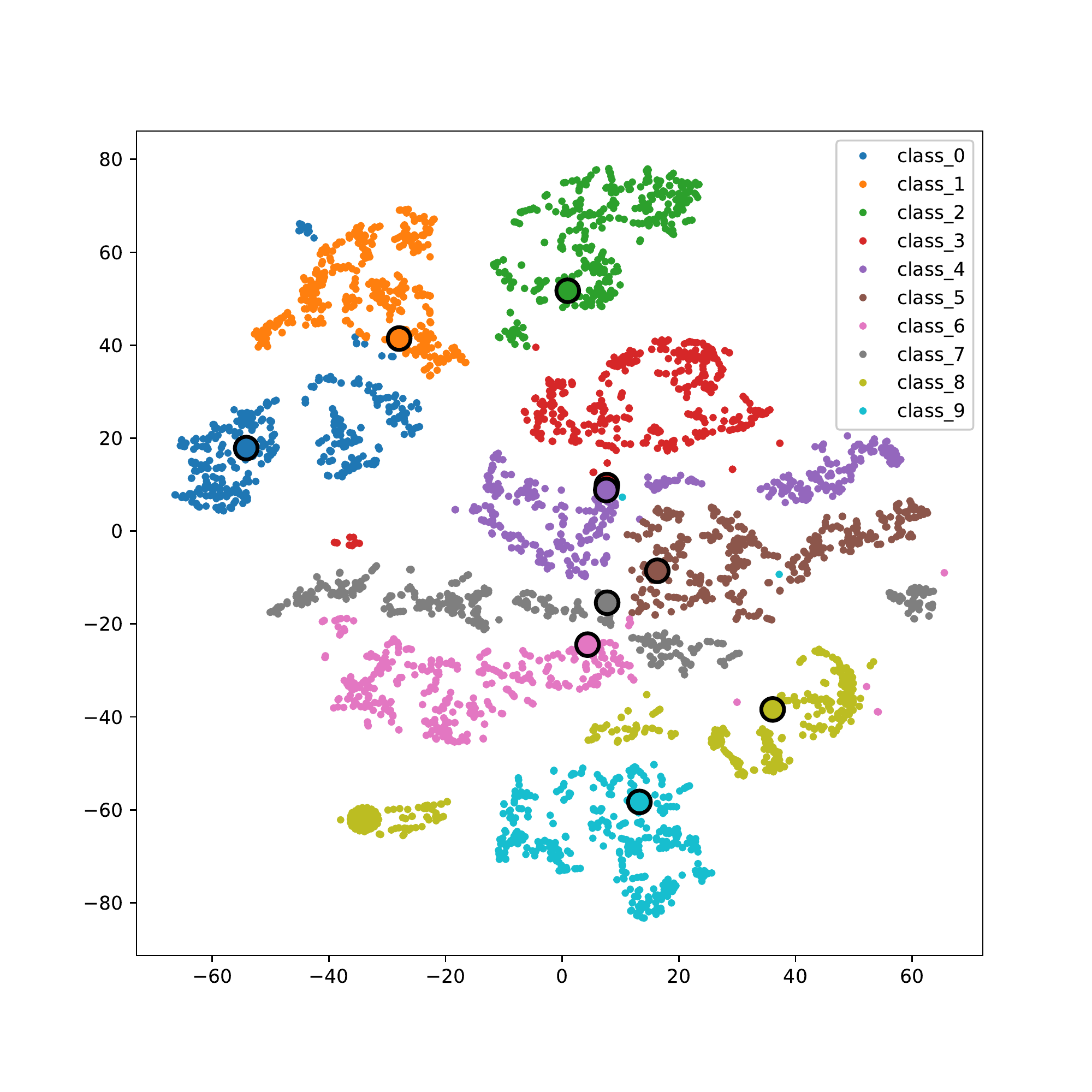}
    \caption{On notMNIST dataset (\textbf{Top}) Reconstruction of images after all tasks have been observed. (\textbf{Bottom}) t-SNE plot of each class after all tasks have been observed.}
    \label{fig:recon_notmnist}
\end{figure}
\begin{figure}[!htpb]
    \centering
    \includegraphics[width = 0.8\linewidth,trim={4em 7em 4em 7em},clip]{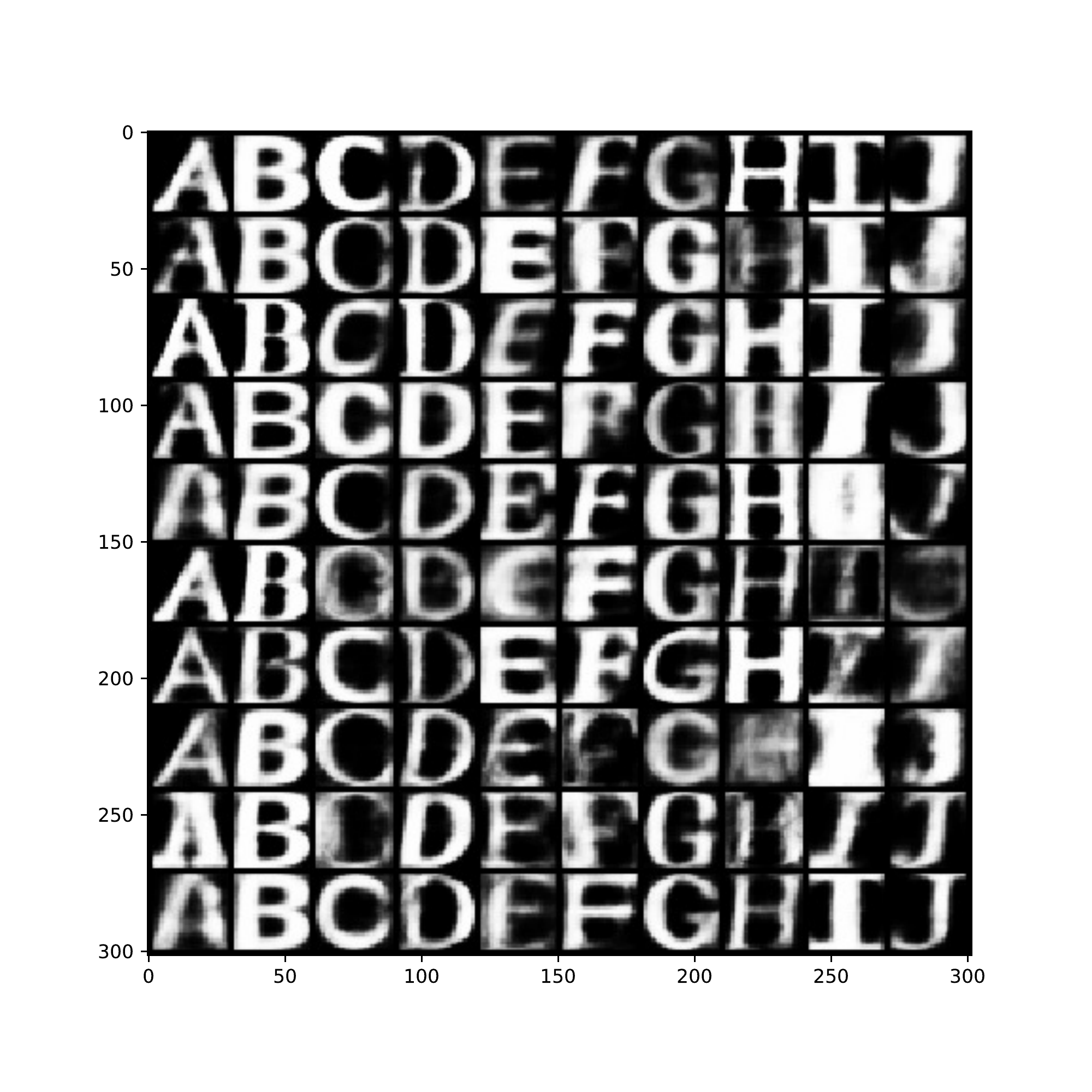}
    \caption{Generated samples on notMNIST dataset after all tasks have been observed}
    \label{fig:gens_notmnist}
\end{figure}
Similarily, Fig \ref{fig:recon_notmnist} shows the reconstructed images of not MNIST dataset and the t-SNE plot of latent codes our model produces, and Fig \ref{fig:gens_notmnist} shows the generated samples from the learned prior over latent space after all tasks are observed. 
\begin{table*}[!htpb]
\centering
  \begin{tabular}{l|l|l|l|l|l|l}
    \hline
    \multirow{2}{*}{\textbf{Benchmarks}} &
      \multicolumn{3}{c}{MNIST} &
      \multicolumn{3}{c}{not MNIST} \\
    & 3-KNN error & 5-KNN error & 10-KNN error &  3-KNN error & 5-KNN error & 10-KNN error \\
    \hline
    Naive & 30.1\% & 33.1\% & 36.0\% & 20.6\% & 24.87\% & 30.8\% \\
    \hline
    EwC & 16.6\% & 19.48\% & 22.3\% & 11.7\% & 13.1\% & 17.8\% \\
    \hline
    VCL & 17.0\% & 19.02\% & 30.2\% & 12.3\% & 13.8\% & 16.5\% \\
    \hline
    Ours & \textbf{0.37}\% & 0.40\% & 0.51\% & \textbf{0.08}\% & 0.09\% & 0.21\% \\
    \hline
  \end{tabular}
  \caption{Unsupervised learning benchmark comparison with sampled latents using K-nearest neighbour test.}
  \label{KNN}
\end{table*}
\subparagraph{Representation Learning} In t-SNE plots, it can be observed that the latent space for MNIST dataset is more clearly seperated as compared to notMNIST dataset. This can be attributed to the abundance of data and less variation in MNIST dataset as compared to notMNIST dataset. we further analyzed the representations that were learned by our model by doing $K$-Nearest Neighbour classification on the latent space. Table \ref{KNN} shows the KNN test error of our model and few other benchmarks on MNIST and notMNIST datasets. We performed the test with three different values for $K$. As shown in the table, the representations learned by other baselines are not very useful (as evidenced by the large test errors), since the latent space are not shared among the tasks, whereas our model uses a shared latent space (yet modulated for each task based on the learned task-specific mask) which results in effective latent representation learning. 
\begin{figure}[!tb]
    \centering
    \includegraphics[width = 0.75\linewidth,trim={4em 5em 4em 5em},clip]{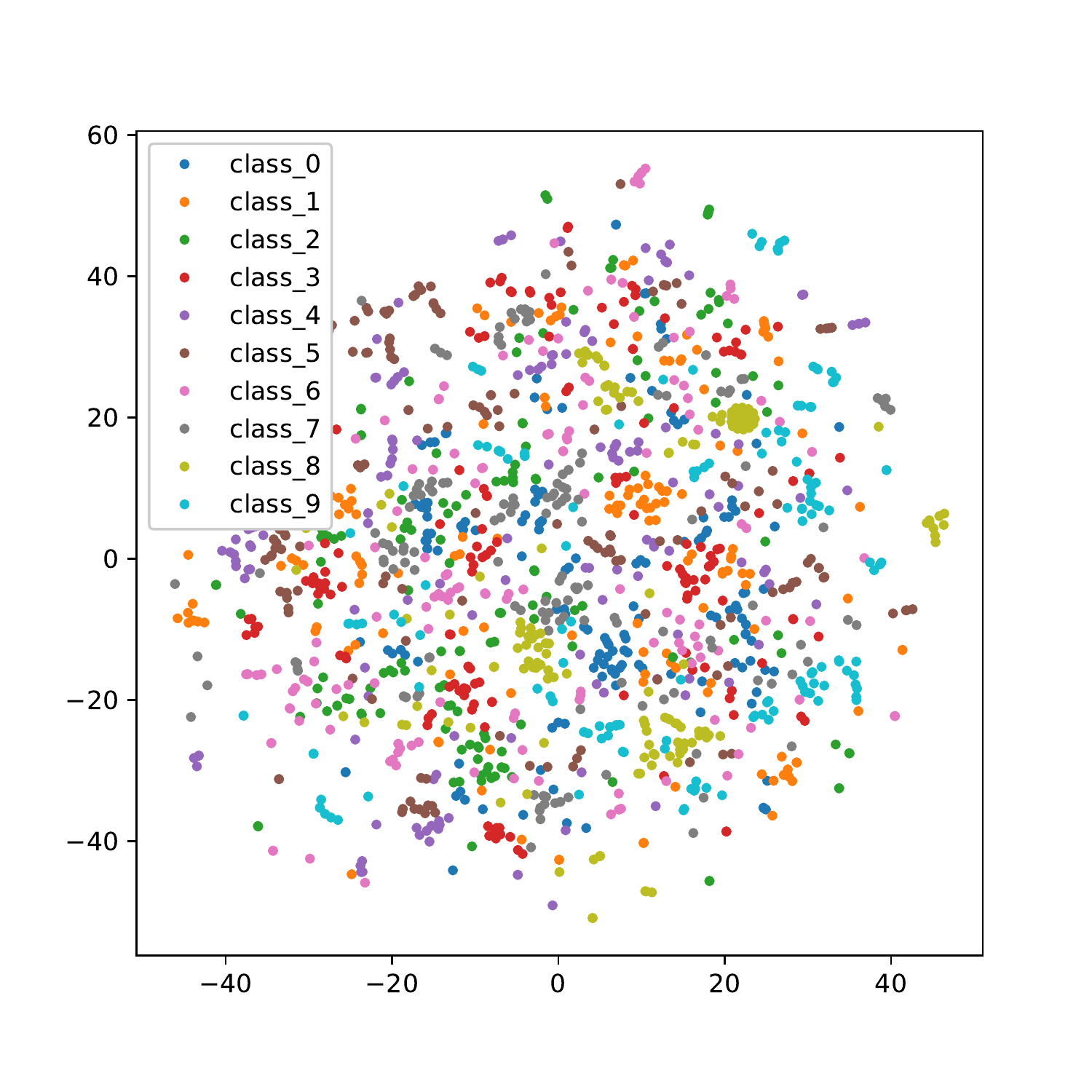}
    \includegraphics[width = 0.75\linewidth,trim={4em 5em 4em 5em},clip]{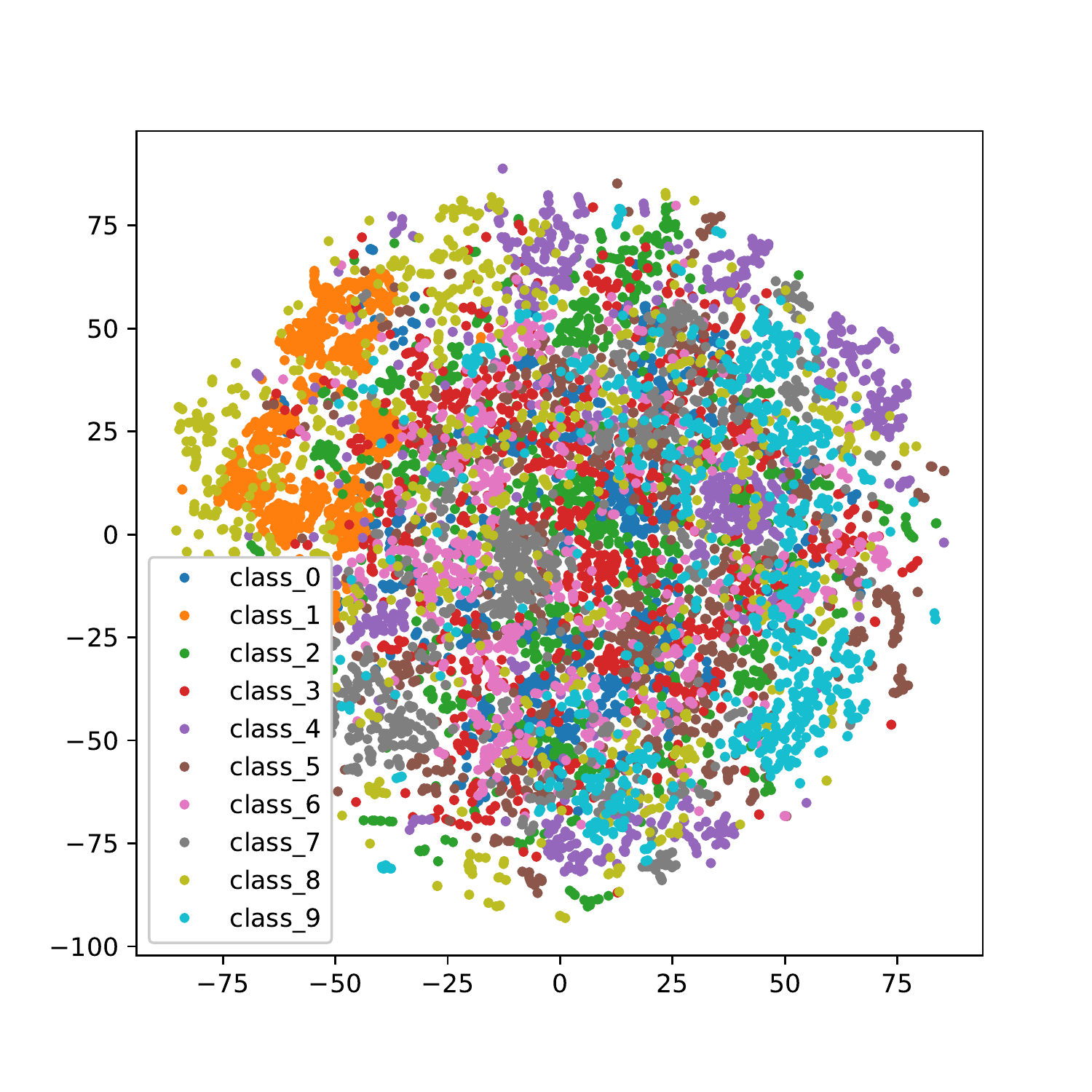}
    \caption{t-SNE plot of latent space of VCL model on notMNIST (top)  and MNIST (bottom) datasets}
    \label{fig:t_SNE_vcl}
\end{figure}

\bibliography{references}
\bibliographystyle{icml2020}

\end{document}